\documentclass[letterpaper]{article} 

\usepackage{fbox}
\usepackage{amsmath}
\usepackage{amsthm}
\newtheorem{theorem}{Theorem}
\usepackage{amsfonts}
\usepackage{subcaption}
\usepackage{booktabs}
\usepackage{multirow}
\usepackage{tabularx}
\usepackage{amssymb}
\usepackage{makecell}
\usepackage{bm}

\usepackage{aaai2026}  
\usepackage{times}  
\usepackage{helvet}  
\usepackage{courier}  
\usepackage[hyphens]{url}  
\usepackage{graphicx} 
\def\UrlFont{\rm}  
\usepackage{natbib}  
\usepackage{caption} 
\usepackage{algorithm}
\usepackage{algorithmic}
\newcolumntype{Y}{>{\centering\arraybackslash}X}
\usepackage{newfloat}
\usepackage{listings}
\DeclareCaptionStyle{ruled}{labelfont=normalfont,labelsep=colon,strut=off} 
\floatstyle{ruled}
\newfloat{listing}{tb}{lst}{}
\floatname{listing}{Listing}
\title{Transolver Is a Linear Transformer: Revisiting Physics-Attention Through the Lens of Linear Attention}
\author{
    Wenjie Hu\textsuperscript{\rm 1,2}\equalcontrib,
    Sidun Liu\textsuperscript{\rm 1,2}\equalcontrib,
    Peng Qiao\textsuperscript{\rm 1,2}\thanks{Co-corresponding author.},
    Zhenglun Sun\textsuperscript{\rm 1,2},
    Yong Dou\textsuperscript{\rm 1,2}\footnotemark[2]
}
\affiliations{
    \textsuperscript{\rm 1}College of Computer Science and Technology, 
    National University of Defense Technology,
    Changsha,
    China\\
    \textsuperscript{\rm 2}National Key Laboratory of Parallel and Distributed Computing,
    National University of Defense Technology,
    Changsha,
    China\\

    \{hwjb127,liusidun,pengqiao,zhenglun\_sun,yongdou\}@nudt.edu.cn
}

\usepackage{bibentry}

\begin{document}

\maketitle
\begin{abstract}
Recent advances in Transformer-based Neural Operators have enabled significant progress in data-driven solvers for Partial Differential Equations (PDEs). Most current research has focused on reducing the quadratic complexity of attention  to address the resulting low training and inference efficiency. 
Among these works, Transolver stands out as a representative method that introduces Physics-Attention to reduce computational costs. Physics-Attention projects grid points into slices for slice attention, then maps them back through deslicing.
However, we observe that Physics-Attention can be reformulated as a special case of linear attention, and that the slice attention may even hurt the model performance.
Based on these observations, we argue that its effectiveness primarily arises from the slice and deslice operations rather than interactions between slices. 
Building on this insight, we propose a two-step transformation to redesign Physics-Attention into a canonical linear attention, which we call \textit{Linear Attention Neural Operator} (LinearNO).
Our method achieves state-of-the-art performance on six standard PDE benchmarks, while reducing the number of parameters by an average of 40.0\% and computational cost by 36.2\%. Additionally, it delivers superior performance on two challenging, industrial-level datasets: AirfRANS and Shape-Net Car.
\end{abstract}

\begin{links}
    \link{Code}{https://github.com/HiPRL/LinearNO}
\end{links}

\section{Introduction}

Solving Partial Differential Equations is a fundamental task in many fields of science and engineering. Due to their complexity, these equations often require discretization into meshes and solving with numerical methods~\cite{piomelli1999large,alfonsi2009reynolds,lee2015direct}, which are computationally expensive and time-consuming. Recent advances in deep learning provide new approaches for PDE solving~\cite{chen2022flowdnn,liu2024lkflownet,huang2025partial}.

The essence of PDE solving is to find the solution function corresponding to the PDE and its boundary or initial conditions. This can be formulated as a supervised learning problem, where neural networks learn the mappings from boundary or initial functions to solution functions. Neural Operator~\cite{luLearningNonlinearOperators2021,kovachki2023neural} was proposed to learn the mappings between those two function spaces while guaranteeing both discretization-invariance and universal approximation. 

\begin{figure}[t]
    \centering
    \includegraphics[width=.75\linewidth]{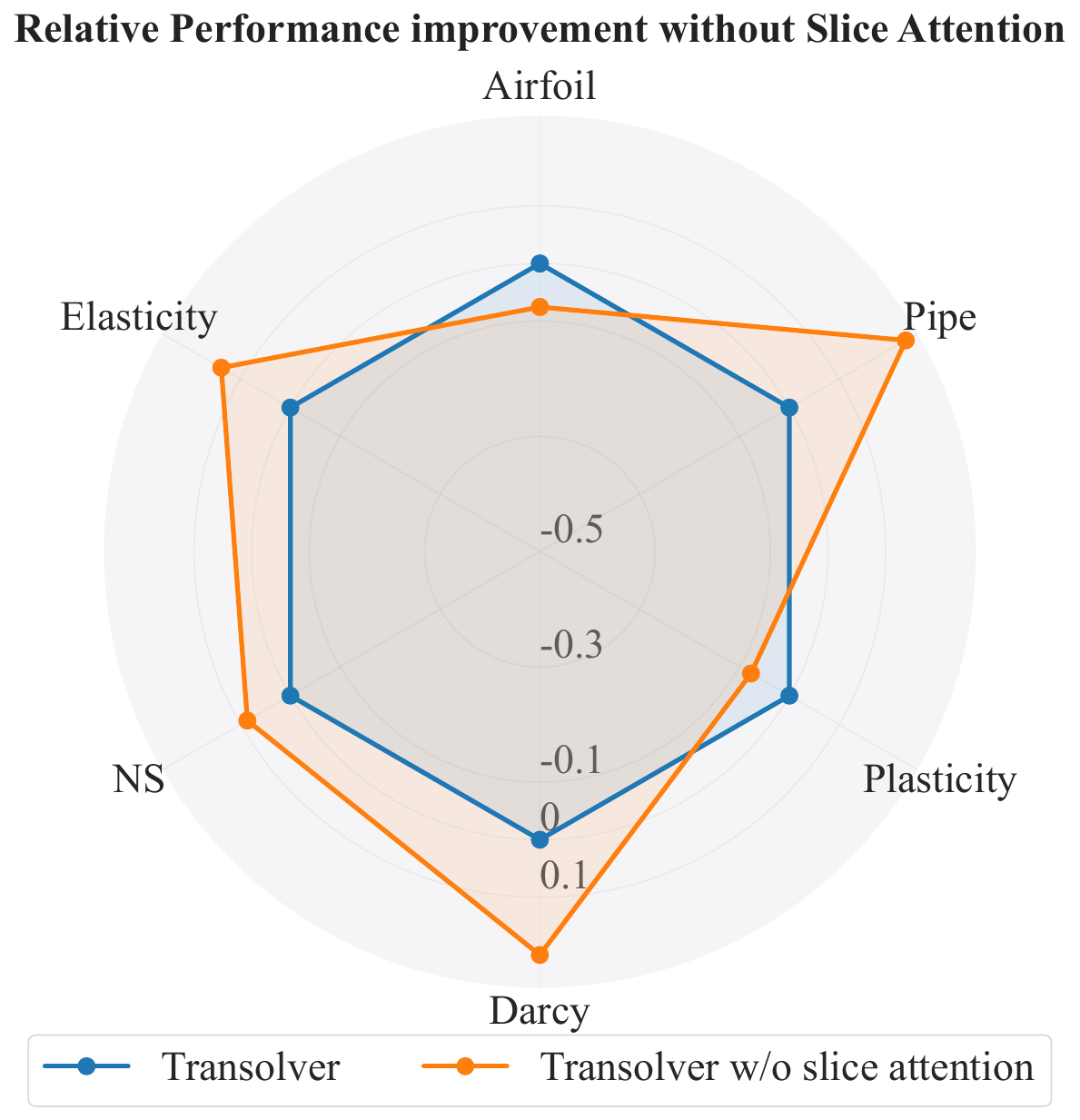}
    \caption{We observe that in most scenarios, removing the slice attention in Physics-Attention leads to performance improvement. This suggests that its effectiveness mainly stems from the slice and deslice operations, rather than the interactions between slices. }
    \label{fig:radar_plot}
\end{figure}
The integral kernel operator of Neural Operator~\cite{kovachki2023neural} is a non-local operation that passes messages between discrete grid points in a discretization-invariant manner, which is analogous to the transformer's self-attention~\cite{vaswani2017attention}. Therefore, a series of works have introduced transformers into PDE solving and Neural Operator construction. However, when considering discrete grid points as tokens, the quadratic complexity of self-attention seriously limits the scale of PDE problems. Several strategies are employed to optimize the complexity of self-attention. 
Some methods~\cite{li2023transformer,li2023scalable,haoGNOTGeneralNeural2023} introduce linear attention~~\cite{DBLP:conf/icml/KatharopoulosV020} to decrease the quadratic complexity to linear. Other methods~\cite{wangLatentNeuralOperator2024,serranoAROMAPreservingSpatial2024,alkinUniversalPhysicsTransformers2024} follow the Set Transformer~\cite{lee2019set}, projecting grid points into a latent space to reduce their length and performing attention in that space for PDE solving. All of the above works aim to construct low-rank patterns to achieve complexity reduction. Transolver~\cite{wu2024transolver}, as a representative model, proposes a Physics-Attention module which contains a slicing operation to project the grid points into several slices and perform slice attention, and a deslicing operation to project the slices back to grid points. However, the design of this module is still intuitive, and lacks in-depth analysis. 

In this paper, we reveal that the Transolver's Physics-Attention is actually a special case of linear attention. 
We further conduct preliminary experiments and observe that in most scenarios, removing the slice attention in Physics-Attention leads to performance improvement, as shown in Figure~\ref{fig:radar_plot}.
This suggests that the effectiveness of Physics-Attention mainly comes from the slice and deslice operations rather than interactions between slices. Based on this insight, we conduct a two-step transformation to make Physics-Attention a canonical linear attention. Specifically, in the generalization step, we relax certain constraints of the Physics-Attention to align it with linear attention. In the simplification step, we demonstrate the non-essential nature of slice attention and consequently remove it.
The derived \textit{Linear Attention Neural Operator}, or LinearNO, retains a canonical structure and achieves higher PDE solving accuracy than Physics-Attention with fewer parameters and computational cost. We further prove that LinearNO is a Monte Carlo approximation of the integral kernel operator.

Our main contributions can be summarized as follows:
\begin{itemize}
\item We reveal that the Physics-Attention proposed by Transolver is essentially a special case of linear attention.
\item We propose a two-step transformation to make Physics-Attention a canonical linear attention, and illustrate the rationale behind it.
\item The derived linear attention model outperforms Transolver with fewer parameters and computational cost, achieving state-of-the-art on multiple PDE solving tasks.
\end{itemize}

\begin{figure*}[t]
    \centering
    \hfill
    \begin{subfigure}[b]{0.3\textwidth}
        \centering
        \includegraphics[width=\linewidth]{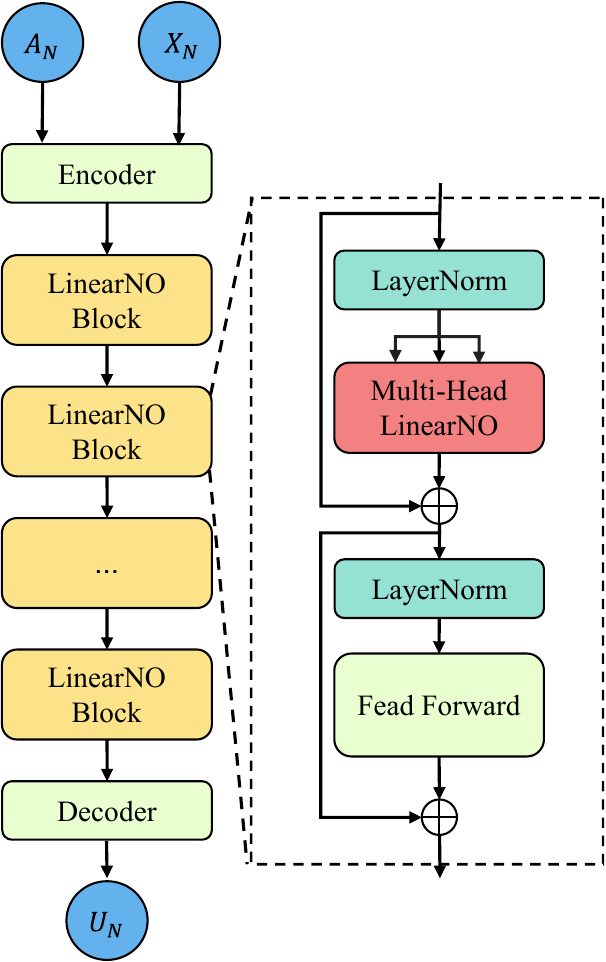}
        \caption{The overall design}
        \label{fig:Pipeline}
    \end{subfigure}
    \begin{subfigure}[b]{0.65\textwidth}
        \includegraphics[width=\linewidth]{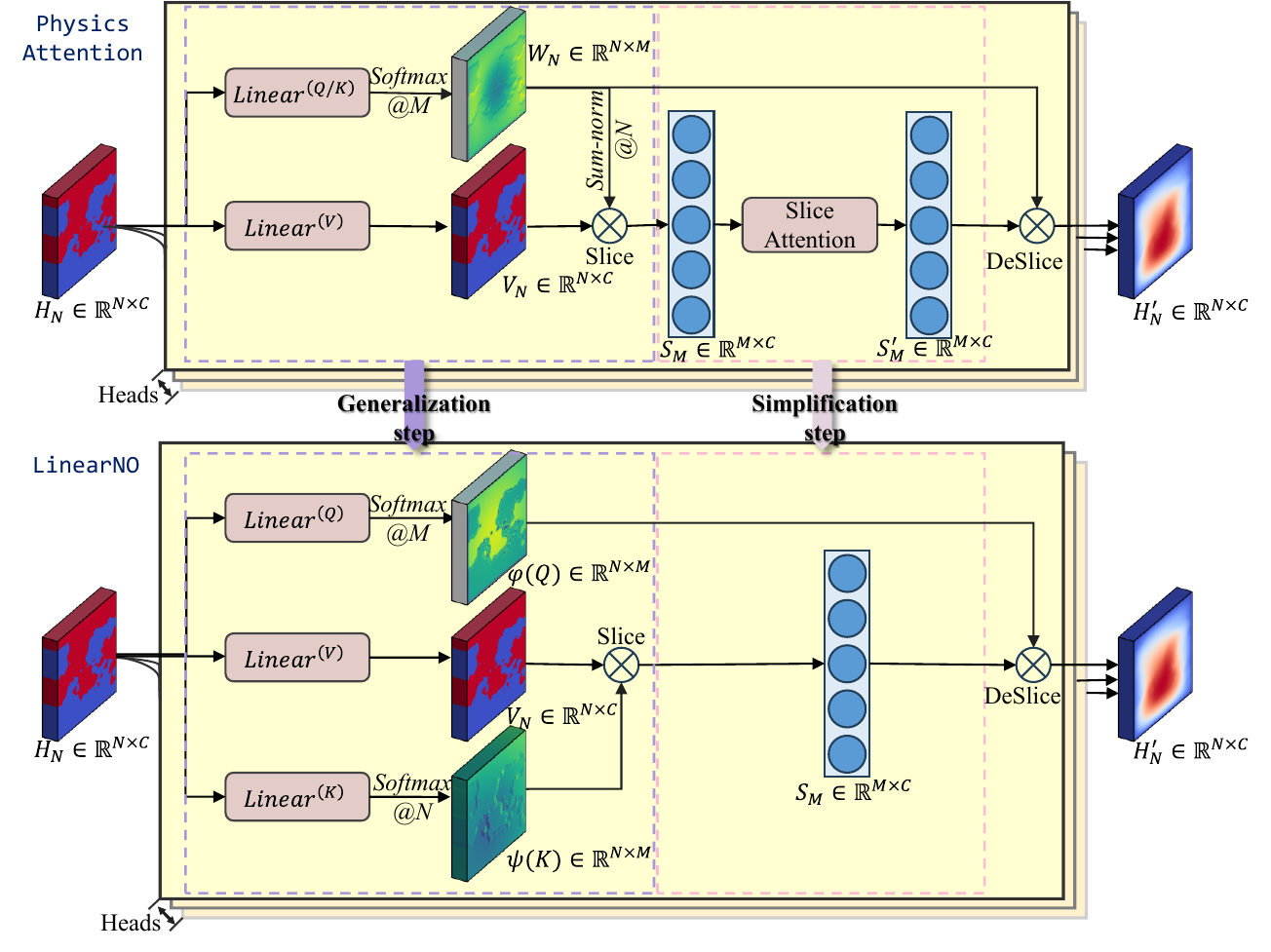}
        \caption{Physics-Attention vs. LinearNO}
        \label{fig:compare}
    \end{subfigure}
    \hfill
    \caption{(a) The overall design of our network. The encoder and decoder modules follow the same architectural design as those in Transolver. (b) Comparison between Physics-Attention and LinearNO. The top shows the Physics-Attention in Transolver, while the bottom depicts our LinearNO.
\textit{Softmax@M} and \textit{Softmax@N} indicate softmax operations along dimensions $M$ and $N$, respectively. \textit{Sum-norm@N} refers to the standard normalization $x_i'=\frac{x_i}{\sum_{j=1}^N x_j}$ for each row.}
    \label{fig:all}
\end{figure*}

\section{Related Work}
\subsection{Neural Operators}

Neural Operators are a class of models that aim to solve PDE by learning a mapping from boundary or initial functions to solution functions. They achieve mappings between function spaces by ensuring universal approximation and discretization-invariance~\cite{kovachki2023neural}. DeepONet~\cite{luLearningNonlinearOperators2021} uses a two-branch architecture to learn the mapping between functions. 
Graph Neural Operator (GNO)~\cite{liNeuralOperatorGraph2020} implements the integral kernel operator using message passing in graph neural networks.
Fourier Neural Operator (FNO)~\cite{liFourierNeuralOperator2021} performs the integral kernel operator by mapping input functions to the frequency domain via Fourier transforms.
GEO-FNO~\cite{DBLP:journals/jmlr/LiHLA23} extends FNO to unstructured meshes by mapping irregular domains onto structured grids.
U-FNO~\cite{DBLP:journals/corr/abs-2109-03697} enhances the multi-scale feature extraction capability of FNO by incorporating a U-Net architecture.

\subsection{Transformers for PDE Solving}
Transformer~\cite{vaswani2017attention} has been shown to be a special case of integral kernel operators~\cite{kovachki2023neural}.
However, its quadratic computational complexity makes it difficult to apply to large-scale PDE problems efficiently.
Some methods, such as OFormer~\cite{li2023transformer}, FactFormer~\cite{li2023scalable}, and GNOT~\cite{haoGNOTGeneralNeural2023}, adopt  linear attention to reduce the computational cost while maintaining the modeling capacity.
Another line of methods aims to reduce complexity by compressing the length of the token sequence.
Methods such as LNO~\cite{wangLatentNeuralOperator2024}, AROMAP~\cite{serranoAROMAPreservingSpatial2024}, UPT~\cite{alkinUniversalPhysicsTransformers2024}, and AeroGTO~\cite{liuAeroGTOEfficientGraphTransformer} reduce computational complexity by introducing a learnable set of latent tokens to project discrete grid points into a latent space using cross-attention, followed by self-attention within the latent space.
LSM~\cite{wuSolvingHighDimensionalPDEs2023} also employs the same compression strategy to reduce sequence length, and replaces self-attention with spectral transformation for compressed tokens.
Transolver~\cite{wu2024transolver} introduces Physics-Attention, which uses a learnable slice-weight matrix to assign discrete grid points to physical slices, grouping similar grid points together. Self-attention is then applied across slices to exchange information and uncover deeper physical relationships within the complex geometry. The slice-weight matrix is used to decode the slices back to grid points. Transolver++~\cite{DBLP:journals/corr/abs-2502-02414} proposes local adaptation for better slice distinction.

\subsection{Linear Attention}
The Softmax operation on the attention matrix is the key factor contributing to the quadratic complexity of attention. To address this issue, linear attention uses kernel functions to approximate the Softmax function, allowing it to change the computation order of $\bm{Q}$, $\bm{K}$, and $\bm{V}$ in full attention, thereby reducing the computational complexity from quadratic to linear.
The Linear Transformer~\cite{DBLP:conf/icml/KatharopoulosV020} replaces the softmax operation with a kernel function, achieving performance comparable to standard attention. Both Performer~\cite{DBLP:journals/corr/abs-2006-03555} and RFA ~\cite{DBLP:conf/iclr/Peng0Y0SK21} employ random feature methods to approximate the softmax function, making the approximation mathematically equivalent to full attention. TSSA~\cite{DBLP:conf/iclr/WuDLPZWY0H25} derives a new linear attention by optimizing a variational form of the Maximum Coding Rate Reduction objective.
RALA~\cite{fan2025breaking} introduces a rank-augmented method to enhance the performance.

\section{Methodology}
\subsection{Preliminaries}

\subsubsection{Problem Setting.}
For a PDE problem defined on the spatial domain $\Omega \subseteq \mathbb{R}^d$, the domain $\Omega$ is discretized into $N$ grid points, denoted as $\{\bm{x}_i\}_{i=1}^{N}=\bm{X}_N \in \mathbb{R}^{N\times d}$, where $\bm{X}_N$ contains the coordinates of all grid points. Correspondingly, we obtain the discretized form of the initial/boundary condition function, $\{a(\bm{x}_i)\}_{i=1}^{N}=\bm{A}_N \in \mathbb{R}^{N\times d_a}$, and the discretized solution, $\{u(\bm{x}_i)\}_{i=1}^{N}=\bm{U}_N \in \mathbb{R}^{N\times d_u}$. Our objective is to learn the mapping from the coordinates and initial/boundary condition function $(\bm{X}_N,\bm{A}_N)$ to the corresponding solution function $\bm{U}_N$, as governed by the operator $L_a$. The inclusion of $\bm{A}_N$ is problem-dependent.

\subsubsection{Transolver Revisiting.}
Transolver \cite{wu2024transolver} proposes a Transformer-based Neural Operator to capture the physical properties of PDE solutions through the Physics-Attention. In this section, we briefly revisit its core components.
The input to the Physics-Attention is a feature matrix $\bm{H}_N=\{\bm{h}_i\}_{i=1}^{N} \in \mathbb{R}^{N\times d_h}$, obtained from $\bm{A}_N$ and $\bm{X}_N$, where $N$ is the number of spatial grid points and $d_h$ is the feature dimension. First, the input feature matrix $\bm{H}_N$ is passed through a linear layer followed by a softmax function to produce a slice-weight matrix $\bm{W}_N=\{\bm{w}_i\}_{i=1}^{N} \in \mathbb{R}^{N\times M}$, where $M$ is a hyperparameter denoting the number of slices.
A channel-wise linear layer is subsequently applied to $\bm{H}_N$ to obtain $\bm{V}_N=\{\bm{v}_i\}_{i=1}^{N} \in \mathbb{R}^{N\times d_h}$.
Using this slice-weight matrix, Physics-Attention computes a weighted average of $\bm{V}_N$ to obtain a slice feature matrix $\bm{S}_M=\{\bm{s}_j\}_{j=1}^{M} \in \mathbb{R}^{M\times d_h}$. This operation can be interpreted as projecting the original $N$ grid points onto $M$ physical slices, effectively compressing the sequence length from $N$ to $M$. The computation is formally given by:
\begin{equation}
    \bm{w}_i=\text{Softmax}(\text{Linear}^{(Q/K)}(\bm{h}_i)) \quad i=1,\dots,N
    \label{eq:weight}
\end{equation}
\begin{equation}
    \bm{v}_i=\text{Linear}^{(V)}(\bm{h}_i) \quad i=1,\dots,N
    \label{eq:v}
\end{equation}
\begin{equation}
    \bm{s}_j=\frac{\sum_{i=1}^{N}\bm{w}_{ij}\bm{v}_i}{\sum_{i=1}^{N}\bm{w}_{ij}} \quad j=1,\dots,M
    \label{eq:slice}
\end{equation}
The $\text{Linear}^{(Q/K)}(\cdot)$ and $\text{Linear}^{(V)}(\cdot)$ are single-layer MLPs with output dimensions of $M$ and $d_h$, respectively.
A self-attention on the slices $\bm{S}_M$ is then performed to produce an updated slices $\bm{S'}_M=\{\bm{s'}_j\}_{j=1}^{M} \in \mathbb{R}^{M\times d_h}$. Finally, using the slice-weight matrix $\bm{W}_N$, the slices $\bm{S'}_M$ are decoded back to the original sequence length, yielding the output $\bm{H'}_N=\{\bm{h'}_i\}_{i=1}^{N} \in \mathbb{R}^{N\times d_h}$.
The process is summarized as follows:
\begin{equation}
    \bm{S'}_M=\text{Self-Attention}(\bm{S}_M)
    \label{eq:attention}
\end{equation}
\begin{equation}
    \bm{h'}_i=\sum_{j=1}^{M}\bm{w}_{ij}\bm{s'}_j \quad i=1,\dots,N
    \label{eq:deslice}
\end{equation}

\subsubsection{Linear Attention.}
As shown in Eq.~\eqref{eq:linearattention}, linear attention aims to approximate the softmax function by designing different kernel functions $\varphi(\cdot)$ and $\psi(\cdot)$, allowing it to change the computation order of $\bm{Q}$, $\bm{K}$, and $\bm{V}$.
In OFormer~\shortcite{li2023transformer} and FactFormer~\shortcite{li2023scalable}, the kernel functions correspond to vector normalization and the identity function, respectively.
The choice of kernel functions directly affects the performance of linear attention.
\begin{equation}
    \begin{aligned}
        \text{Attention}(\bm{Q},\bm{K},\bm{V})&=\text{Softmax}(\frac{\bm{QK}^\top}{\sqrt{d}})\bm{V} \\
        & \approx \varphi(\bm{Q}) (\psi^\top(\bm{K}) \bm{V}) 
        \label{eq:linearattention}
    \end{aligned}
\end{equation}

Some improvements of linear attention insert additional intermediate operations between $\psi^\top(\bm{K}) \bm{V}$ and $\varphi(\bm{Q})$, i.e., $\varphi(\bm{Q}) \circ \mathcal{G} \circ \psi^\top(\bm{K}) \bm{V}$. The operation $\mathcal{G}$ can be rank augmentation~\cite{fan2025breaking}, forget gate~\cite{yang2023gated}, etc.

\begin{figure}[t]
    \centering
    \includegraphics[clip,trim=1mm 1mm 1mm 1mm, width=.7\linewidth]{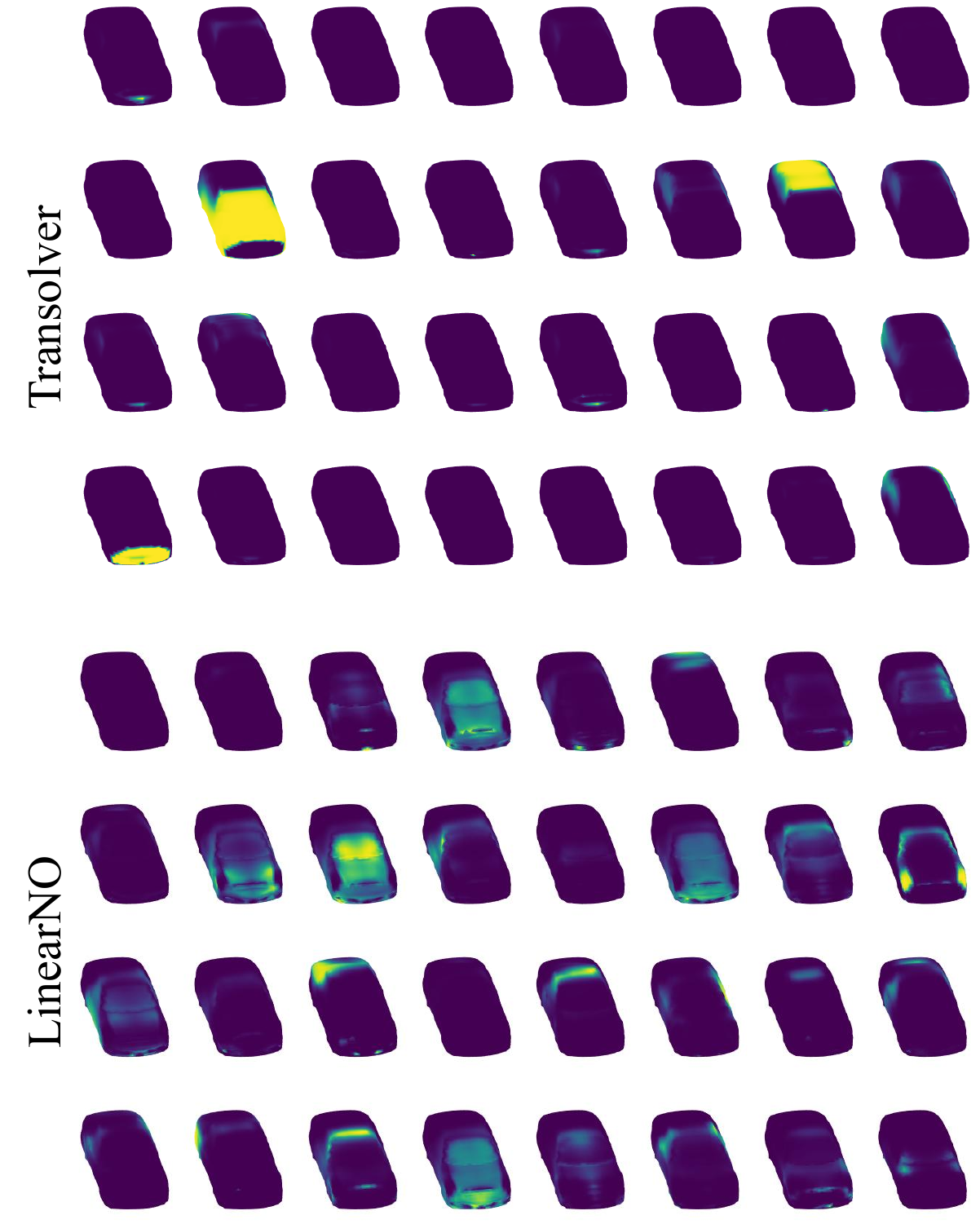}
    \caption{Visualization of Slice-Weight Matrix $W_N$}
    \label{fig:car_slice}
\end{figure}

\subsection{Equivalence of Transolver to Linear Attention}
By comparing the two definitions, we observe that Physics-Attention is essentially a special case of linear attention. 
Specifically, we can construct equivalent feature mappings $\varphi(\bm{Q})$, $\psi(\bm{K})$, and $\bm{V}$ as in Eq.\eqref{eq:equivalence},
\begin{equation}
\label{eq:equivalence}
    \begin{aligned}
    \varphi(\bm{Q})&=\left\{\text{Softmax}(\text{Linear}^{(Q/K)}(\bm{h}_i))\right\}_{i=1}^{N}=\left\{\bm{w}_i\right\}_{i=1}^{N}    \\
    \psi(\bm{K})&=\left\{\frac{\bm{w}_i}{\sum_{j=1}^{N}\bm{w}_j}\right\}_{i=1}^{N} \\
    \bm{V}&=\left\{\text{Linear}^{(V)}(\bm{h}_i)\right\}_{i=1}^{N}    \\    
    \end{aligned}
\end{equation}
and cast the intermediate operation $\mathcal{G}$ as self-attention to reformulate Physics-Attention into a linear attention form.
The linear attention term $\varphi(\bm{Q})$ is equivalent to the slice-weight matrix $\bm{W}_N$ used in Physics-Attention.
In the following sections, we will use these two notations interchangeably depending on the context.

As a result, Physics-Attention is a special case of linear attention where (1) $\varphi(\bm{Q})$ and $\psi(\bm{K})$ are from the same learnable layer $\text{Linear}^{(Q/K)}(\cdot)$ and differ only in their respective normalization schemes, and (2) self-attention is applied as the intermediate operation $\mathcal{G}$. However, the above two characteristics do not help the model performance, which we will discuss further later.

\subsection{LinearNO: Linear Attention Neural Operator}
In this section, we present a more generalized analysis of the Physics-Attention.
Based on this, we redesign it through two key steps—generalization and simplification—to derive a more flexible architecture called the \textit{Linear Attention Neural Operator}, or LinearNO.
Figure~\ref{fig:compare} shows a comparison between the proposed LinearNO and the Physics-Attention.

\subsubsection{Generalization step.}  \citeauthor{DBLP:journals/corr/abs-2502-02414}~\shortcite{DBLP:journals/corr/abs-2502-02414} observed that Physics-Attention tends to produce uniform slice weights, thus generating less distinguishable slices in some cases, which damages the model's performance, as shown in Figure~\ref{fig:car_slice}. We hypothesize that this may be attributed to the intended asymmetry between $\varphi(\bm{Q})$ and $\psi(\bm{K})$ in linear attention; enforcing the shared learnable layer between $\varphi(\bm{Q})$ and $\psi(\bm{K})$ could blur their roles and result in overly averaged representations. Based on this hypothesis, we relaxed the weight-sharing constraint in the generalization step, resulting in asymmetric $\varphi(\bm{Q})$ and $\psi(\bm{K})$. Specifically, we allow $\psi(\bm{K})$ to be learned independently of $\varphi(\bm{Q})$. Following Physics-Attention, $\varphi(\bm{Q})$ and $\psi(\bm{K})$ are normalized along the $M$ and $N$ dimensions, respectively. This ensures that each row of attention matrix $\varphi(\bm{Q})\psi^\top(\bm{K})$ is normalized, which is consistent with full attention~\cite{shen2021efficient}. Formally, we define $\varphi(\bm{Q})$ and $\psi(\bm{K})$ as follows:

\begin{equation}
    \begin{aligned}
        \varphi(\bm{Q}) &= \text{Softmax}(\text{Linear}^{(Q)}(\bm{H}_N)) \\
        \psi^\top(\bm{K}) &= \text{Softmax}(\text{Linear}^{(K)}(\bm{H}_N)^\top)
        \label{eq:phi}
    \end{aligned}
\end{equation}
where $\bm{H}_N\in \mathbb{R}^{N\times d}$ is the input feature, $\text{Linear}^{(Q)}(\cdot)$ and $\text{Linear}^{(K)}(\cdot)$ both contain learnable parameters of shape $\mathbb{R}^{d_h \times M}$, and Softmax is computed row by row. After this step, we observed more diverse attention patterns and a more saturated rank in the attention matrix.

\begin{table*}[htbp]
    \centering
    
    \begin{tabularx}{\linewidth}{l|YYYYYY}
        \toprule
        Model               & Airfoil   & Pipe      & Plasticity      & NS        & Darcy     & Elasticity \\
        \midrule
        FNO \shortcite{liFourierNeuralOperator2021}                 & /         &/          &/          &0.1556     &0.0108     &/     \\
        U-FNO \shortcite{DBLP:journals/corr/abs-2109-03697}        &0.0269     &0.0056     &0.0039     &0.2231     &0.0183     &0.0239 \\
        GEO-FNO \shortcite{DBLP:journals/jmlr/LiHLA23}             &0.0138     &0.0067     &0.0074     &0.1556     &0.0108     &0.0229 \\
        F-FNO \shortcite{DBLP:conf/iclr/TranMXO23}  &0.0078     &0.0070     &0.0047     &0.2322     &0.0077     &0.0263 \\
        \midrule
        Galerkin \shortcite{DBLP:conf/nips/Cao21}            &0.0118     &0.0098     &0.0120     &0.1401     &0.0084     &0.0240\\
        OFormer \shortcite{li2023transformer}             &0.0183     &0.0168     &0.0017     &0.1705     &0.0124     &0.0183\\
        GNOT \shortcite{haoGNOTGeneralNeural2023}                &0.0076     &0.0047     &0.0336     &0.1380     &0.0105     &0.0086\\
        FactFormer \shortcite{li2023scalable}          &0.0071     &0.0060     &0.0312     &0.1214     &0.0109     &/   \\
        ONO \shortcite{DBLP:conf/icml/XiaoHLD024}                 &0.0061     &0.0052     &0.0048     &0.1195     &0.0076     &0.0118\\
        \midrule
        LSM \shortcite{wuSolvingHighDimensionalPDEs2023}                 &0.0059     &0.0050     &0.0025     &0.1535     &0.0065     &0.0218 \\
        LNO$^*$ \shortcite{wangLatentNeuralOperator2024}                &0.0053     &0.0031     &0.0028     &\underline{0.0830}     &0.0063     &0.0066 \\   
        Transolver$^*$ \shortcite{wu2024transolver}          & 0.0053    & 0.0030    & \underline{0.0013}    & 0.0882    & \underline{0.0055}   & 0.0065 \\
        Transolver++$^\dagger$ \shortcite{DBLP:journals/corr/abs-2502-02414} &\underline{0.0051}    &\underline{0.0027} &0.0014     &0.1010 &0.0056 &\underline{0.0064}       \\
        \midrule
        LinearNO (ours)   & \textbf{0.0049}    & \textbf{0.0024}    & \textbf{0.0011}  &\textbf{0.0699}   & \textbf{0.0050} & \textbf{0.0050} \\
        \bottomrule
    \end{tabularx}
    \caption{Comparison of relative L2 errors on six standard PDE benchmark tasks.
    / denotes the method is not applicable to the task.
    $^*$ denotes results reproduced by us. 
    $^\dagger$ denotes results reproduced based on descriptions in the paper due to unavailable source code.
An underscore indicates the second-best result, and bold indicates the best result.}
    \label{tab:StandardPDE}
\end{table*}
\subsubsection{Simplification step.}
In Physics-Attention, the slice ($\psi(\bm{K})$) and deslice ($\varphi(\bm{Q})$) operations are symmetric, which limits the scope of feature interaction to a token itself and those similar to it. As a result, cross-slice feature interaction is not possible, making slice attention necessary. However, after the generalization step lifts the symmetry constraint between $\varphi(\bm{Q})$ and $\psi(\bm{K})$, each token can interact with all others during the slice and deslice processes. This makes the slice attention unnecessary. 
Besides, the experiment in Figure~\ref{fig:radar_plot} also indicates that even under symmetric constraints, slice attention still fails to deliver consistent performance gains.
Therefore, in the simplification step, we remove this attention entirely and set the intermediate operation $\mathcal{G}$ to the identity.

\subsubsection{LinearNO.}
Through the two steps above, we derive a canonical form of linear attention, which we refer to as the \textit{Linear Attention Neural Operator}~(LinearNO). It is formalized as follows:
\begin{equation}
    \text{LinearNO}(\bm{H}_N) = \varphi(\bm{Q})\left(\psi^\top(\bm{K})\cdot \text{Linear}^{(V)}(\bm{H}_N)\right)
\end{equation}

Here we provide a theorem to illustrate that LinearNO is a standard Neural Operator.
\begin{theorem}
Let $\{\bm{x}_i\}_{i=1}^{+\infty}$ be a sequence of refined meshes on $\Omega$ with $\bm{x}_i \thicksim \mu_{\Omega}$, and assume that the function $v(\bm{x})$ is bounded on $\Omega$. As $n \to +\infty$, for any $\epsilon > 0$, the proposed LinearNO converges in probability to a continuous integral kernel operator:

$$\lim_{n \to +\infty}  \Pr \left\{ \frac{1}{N}\| \text{LinearNO}(\bm{x}) -\mathcal{F}(\bm{x}) \| \leq \epsilon \right\}=1$$

$$\mathcal{F}(\bm{x}) := \int_{\Omega} \kappa(v(\bm{x}),v(\bm{y}) )v(\bm{y})\bm{R}   \,d\mu_{\Omega}(\bm{y}) $$

Here, $\bm{R}$ are  learnable parameters, $\mathcal{F}(\bm{x})$ is the continuous integral kernel operator, and the kernel function $\kappa$ of LinearNO is defined as:
$$ \kappa(v(\bm{x}),v(\bm{y}))=\frac{\varphi(v(\bm{x})) \exp(\bm{B}^{\top}v(\bm{y})^{\top})}{\int_{\Omega} \exp{(\bm{B}^{\top}v(\bm{y})^{\top})} \,d\mu_{\Omega}(\bm{y}) } $$

where $\bm{B}$ are learnable parameters, and $\varphi(\cdot)$ denotes the point-wise normalization function as defined in Eq.~\eqref{eq:phi}.
\label{thm:linearNO}
\end{theorem}
This theorem shows that LinearNO satisfies the discretization-invariance of Neural Operators and serves as a Monte Carlo approximation in the form of the continuous integral kernel operator, enabling a mapping from function spaces to function spaces.
We provide a detailed proof of Theorem~\ref{thm:linearNO} in Appendix A.

\begin{table*}[t]
    
    \centering
    \begin{tabularx}{\linewidth}{l|YYYYYYYY}
        \toprule
        \multirow{2}{*}{Model}   &\multicolumn{4}{c}{AirfRANS} &\multicolumn{4}{c}{Shape-Net Car}\\ 
        \cmidrule{2-5} \cmidrule{6-9}
        & Volume $\downarrow$   & Surface $\downarrow$      & $C_L \downarrow$         & $\rho_L \uparrow$  & Volume $\downarrow$   & Surface $ \downarrow$  & $C_D \downarrow$  & $\rho_D \uparrow$\\
        \midrule
        MLP        &0.0081 &0.0200 &0.2108 &0.9932 &0.0512 &0.1304 &0.0307 &0.9496\\
        GraphSage \shortcite{DBLP:conf/nips/HamiltonYL17}  &0.0087 &0.0184 &0.1476 &0.9964 &0.0461 &0.1050 &0.0270 &0.9695\\
        PointNet \shortcite{DBLP:conf/cvpr/QiSMG17}&0.0253 &0.0996 &0.1973 &0.9919 &0.0494 &0.1104 &0.0298 &0.9583\\
        GraphUNet \shortcite{DBLP:conf/icml/GaoJ19}&0.0076 &0.0144 &0.1677 &0.9949 &0.0471 &0.1102 &0.0226 &0.9725\\
        MeshGraphNet \shortcite{DBLP:conf/iclr/PfaffFSB21}&0.0214     &0.0387         &0.2252         &0.9945 &0.0354     &0.0781     &0.0168     &0.9840\\
        \midrule
        GNO \shortcite{liNeuralOperatorGraph2020}         &0.0269     &0.0405         &0.2016         &0.9938 &0.0383     &0.0815     &0.0172     &0.9834\\
        
        GEO-FNO \shortcite{DBLP:journals/jmlr/LiHLA23}     &0.0361     &0.0301         &0.6161         &0.9257 &0.1670     &0.2378     &0.0664     &0.8280\\
        \midrule
        GALERKIN \shortcite{DBLP:conf/nips/Cao21}    &0.0074     &0.0159         &0.2336         &0.9951 &0.0339     &0.0878     &0.0179     &0.9764\\
        GNOT \shortcite{haoGNOTGeneralNeural2023}        &0.0049     &0.0152         &0.1992         &0.9942 &0.0329     &0.0798     &0.0178     &0.9833\\
        GINO \shortcite{DBLP:conf/nips/LiKCLKONS0AA23}        &0.0297     &0.0482         &0.1821         &0.9958 &0.0386     &0.0810     &0.0184     &0.9826\\    
        LNO$^*$ \shortcite{wangLatentNeuralOperator2024}        & 0.0214     &0.0268         &0.1480         &0.9744     &0.0269  & 0.0870 & 0.0174 &0.9781\\
        Transolver$^*$ \shortcite{wu2024transolver}  & \underline{0.0023}    & \underline{0.0085}        & \underline{0.1230}        &\underline{0.9978} &\underline{0.0221} &\underline{0.0785} &\underline{0.0117} &\underline{0.9914}    \\
        Transolver++$^\dagger$ \shortcite{DBLP:journals/corr/abs-2502-02414} &0.0068 &0.0159 &0.1880 &0.9910 &0.0226 &0.0800 &0.0132  &\underline{0.9914}\\
        \midrule
        LinearNO (ours)    & \textbf{0.0011}    & \textbf{0.0077}        & \textbf{0.0491}        &\textbf{0.9992}     &\textbf{0.0194}     & \textbf{0.0754}    &\textbf{0.0106} &\textbf{0.9925}\\
        
        \bottomrule
    \end{tabularx}
    \caption{Performance on AirfRANS and Shape-Net Car.
    Volume and Surface represent the physical fields in the surrounding flow region and on the object surface, respectively.
$C_L$ and $C_D$ denote the lift coefficient and drag coefficient, respectively.
The table reports their relative L2 errors.
Spearman's correlation coefficient $\rho$ is closer to 1 indicating better performance.
$^*$ indicates results reproduced by us.
$^\dagger$ denotes results reproduced based on descriptions in the paper due to unavailable source code.
An underscore indicates the second-best result, and bold indicates the best result.}
    \label{tab:PDEIndustrial}
\end{table*}

\section{Experiments}
\subsection{Experimental Setup}
\textbf{Datasets} We adopt six benchmark datasets, including Airfoil, Pipe, Plasticity, NS, Darcy and Elasticity, which are classical problems in fluid mechanics and solid mechanics. These datasets were introduced by FNO~\cite{liFourierNeuralOperator2021} and Geo-FNO~\cite{DBLP:journals/jmlr/LiHLA23}, and are now widely used as standard benchmarks for evaluating Neural Operators. In addition, we use two industrial-level datasets:
AirfRANS \cite{bonnet2022airfrans} and ShapeNet-Car \cite{DBLP:journals/tog/ShapeCar} to assess the model's performance in aerodynamic shape design. More detailed information about all datasets can be found in Appendix B.

\noindent\textbf{Baselines} We compare our model against several representative baselines, including classical Neural Operator models: FNO~\shortcite{liFourierNeuralOperator2021}, Geo-FNO~\shortcite{DBLP:journals/jmlr/LiHLA23}, F-FNO~\shortcite{DBLP:conf/iclr/TranMXO23} and U-FNO~\shortcite{DBLP:journals/corr/abs-2109-03697},
Transformer-based models: Galerkin~\shortcite{DBLP:conf/nips/Cao21}, OFormer~\shortcite{li2023transformer}, GNOT~\shortcite{haoGNOTGeneralNeural2023}, FactFormer~\shortcite{li2023scalable}, ONO~\shortcite{DBLP:conf/icml/XiaoHLD024}, LSM~\shortcite{wuSolvingHighDimensionalPDEs2023}, LNO~\shortcite{wangLatentNeuralOperator2024}, Transolver~\shortcite{wu2024transolver} and Transolver++~\shortcite{DBLP:journals/corr/abs-2502-02414}, 
classical geometric deep models: PointNet~\shortcite{DBLP:conf/cvpr/QiSMG17}, GraphSage~\shortcite{DBLP:conf/nips/HamiltonYL17} and GraphUNet~\shortcite{DBLP:conf/icml/GaoJ19}, MeshGraphNet~\shortcite{DBLP:conf/iclr/PfaffFSB21}.

\noindent\textbf{Setup} To ensure fair comparisons, we follow the experimental settings of Transolver \cite{wu2024transolver}. 
We report the relative L2 error and Spearman correlation coefficient $\rho$ as the evaluation metric. Detailed experimental settings and hyperparameter configurations are provided in Appendix B.

\begin{figure*}[t]
    \centering
    
    \begin{subfigure}[b]{0.99\linewidth}
        \centering
        \includegraphics[width=\linewidth]{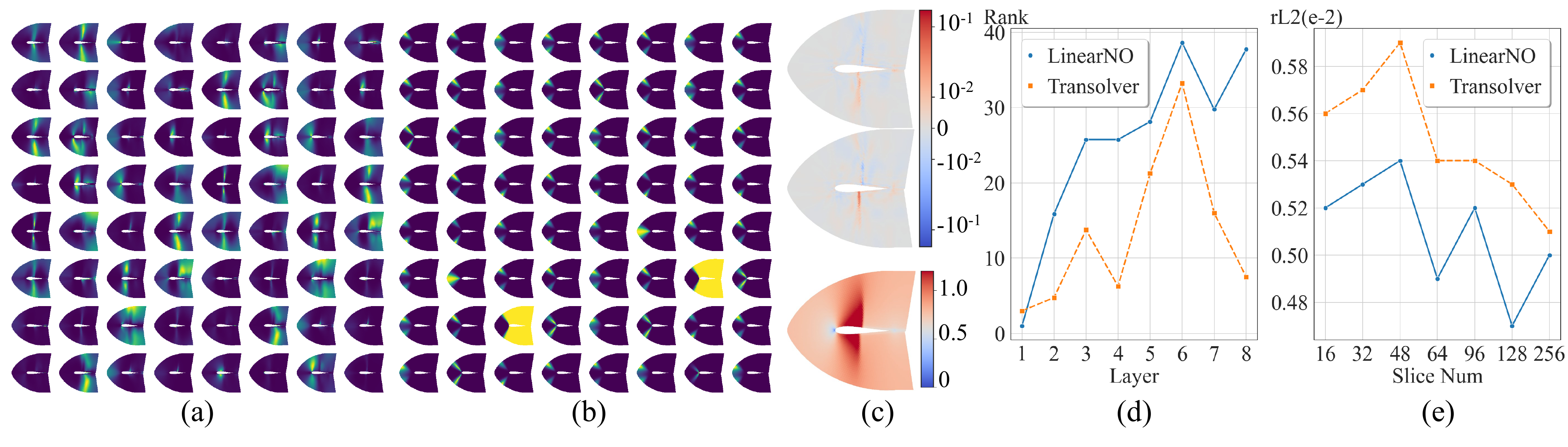}
    \end{subfigure}

    \caption{
    All experiments are conducted on the Airfoil dataset.
    (a) and (b) show the final-layer slice-Weight matrix $W_N$ visualizations of LinearNO and Transolver, respectively.
    (c) shows the error distributions of LinearNO (top) and Transolver (middle), with the ground truth shown at the bottom.
    (d) shows the average rank of the attention matrix $\varphi(\bm{Q})\psi^\top(\bm{K})$ per head at each layer when slice numbers $M = 64$.
    (e) presents the prediction errors of both models under different slice numbers.}
    \label{fig:slice-vis}
\end{figure*}

\subsection{Main Results}
\subsubsection{Accuracy Comparison.}
Table~\ref{tab:StandardPDE} presents the relative L2 errors of LinearNO and SOTA models across six PDE benchmarks. LinearNO achieves the best performance on all tasks. 
Specifically, LinearNO achieves over a 10\% relative improvement on the Pipe, Plasticity, NS and Elasticity task.

We further evaluate our method on two complex, industrial-level high-fidelity datasets: AirfRANS and Shape-Net Car, with results summarized in Table~\ref{tab:PDEIndustrial}.
Our model achieves the best results on both tasks, especially on the AirfRANS dataset, where LinearNO outperforms Transolver by more than 60\% in predicting the lift coefficient $C_L$, achieving a Spearman's correlation coefficient of 0.9992. This demonstrates that our method can more accurately predict the aerodynamic properties of different geometries, thereby assisting engineers in the iterative design of shapes with superior aerodynamic characteristics.
We also conducted generalization experiments on the AirfRANS dataset.
Specifically, the training and testing sets have different ranges of Reynolds numbers and angles of attack.
For more details, please refer to Appendix D.

Overall, these results demonstrate the superior performance of LinearNO and provide strong empirical support for our theoretical generalization and simplification of the Physics-Attention.

\subsubsection{Efficiency Analysis.}
We compare the number of parameters and computational cost between LinearNO and Transolver on six standard PDE tasks, as shown in Table~\ref{tab:Param}. It can be observed that LinearNO has significantly fewer parameters and lower computational cost than Transolver. On average, our method reduces the number of parameters by 40.0\% and the FLOPs by 36.2\%. This demonstrates that LinearNO is more computationally efficient and lightweight, making it more suitable for resource-constrained environments.

\begin{table}[t]
    \centering
    
    \begin{tabular}{c|c|ccc}
        \toprule
        Metric&Model & Airfoil & Darcy &$\text{Elas}^*$ \\
        \midrule
        \multirow{2}{*}{\makecell{Parameter\\(GB)}}  &Transolver&2.81&2.83&0.71\\
                                    &LinearNO&\textbf{1.77}&\textbf{1.77}&\textbf{0.59}\\
        \midrule
        \multirow{2}{*}{\makecell{Computation\\(GFLOPs)}}    &Transolver&32.38&20.87&0.76\\
                                    &LinearNO&\textbf{21.34}&\textbf{13.68}&\textbf{0.69}\\
        \bottomrule
    \end{tabular}
    \caption{Comparison on parameter count and computational cost between LinearNO and Transolver. 
    $\text{Elas}^*$ is used as an abbreviation for Elasticity to fit the table layout. 
    For all benchmark datasets, please refer to Appendix C.}
    \label{tab:Param}
\end{table}
\begin{table}[t]
    \centering
    
    \begin{tabular}{c|cc|ccc}
        \toprule
        No.&\textit{Gen.}  & \textit{Sim.} & Airfoil & Darcy& Elasticity\\
        \midrule
        1&$\times$ & $\times$ & 0.0067& 0.0064& 0.0069\\
        2&$\checkmark$ & $\times$ & 0.0054& 0.0061& 0.0062\\
        3&$\times$ & $\checkmark$ & 0.0071& 0.0052& 0.0064\\
        4&$\checkmark$ & $\checkmark$ & \textbf{0.0049} & \textbf{0.0050}& \textbf{0.0050}\\
        \bottomrule
    \end{tabular}
    \caption{Ablation study of generalization and simplification. \textit{Gen.} denotes the generalization step, and \textit{Sim.} denotes the simplification step. 
    For all benchmark datasets, please refer to Appendix C.}
    \label{tab:Ablation}
\end{table}

\subsubsection{Slice Analysis.}

The rank of attention matrix $\varphi(\bm{Q})\psi^\top(\bm{K})$ corresponds to the number of physical states effectively captured by the model, which influences its performance~\cite{DBLP:journals/corr/abs-2502-02414}. This effective rank is bounded by the number of slices $M$. We hypothesize that the symmetric design of $\varphi(\bm{Q})$ and $\psi(\bm{K})$ in Transolver limits the effective utilization of slices. In this section, we present a quantitative analysis to support this claim.

We estimate the layer-wise rank of the attention matrix using Singular Value Decomposition~\cite{Golub1965SVD}, and the results are shown in Figure~\ref{fig:slice-vis}. The findings indicate that the asymmetric design of $\varphi(\bm{Q})$ and $\psi(\bm{K})$ facilitates more efficient utilization of slices. Furthermore, we visualize the distribution of the final-layer $\varphi(\bm{Q})$ matrix on the Airfoil dataset. Compared to Transolver, LinearNO exhibits more diverse physical slices. In addition, LinearNO consistently outperforms Transolver across different numbers of slices on the Airfoil dataset.

\begin{table}[t]
    \centering
    
    \begin{tabular}{c|ccc}
        \toprule
        Num of Slice $M$ & Airfoil & Darcy &Elasticity\\
        \midrule
        16 &0.0052 &0.0056   &0.0062\\
        32&0.0053&0.0053&0.0056\\
        48&0.0054&0.0051&0.0052\\
        64 &0.0049&0.0050&0.0050\\
        96 &0.0052&\textbf{0.0048}&0.0056\\
        128&\textbf{0.0047}&0.0050&0.0051\\
        256&0.0050&0.0049&\textbf{0.0046}\\
        \bottomrule
    \end{tabular}
    \caption{Relative L2 errors of LinearNO with different numbers of slices $M$. For all
benchmark datasets, please refer to Appendix C.}
    \label{tab:Slice_num}
\end{table}
\begin{table}[t]
    \centering
    \begin{tabular}{cc|ccc}
        \toprule
        $\varphi(Q)$ & $\psi(K)$ & Airfoil & Darcy & Elasticity\\
        \midrule
        $N$ & $M$ &0.0059  &0.0055 &0.0112  \\
        $M$ & $M$ &0.0061  &0.0060  &0.0081  \\
        $N$ & $N$ &0.0068  &0.0056 &0.0095 \\
        $M$ & $N$ &\textbf{0.0049}  & \textbf{0.0050} &\textbf{0.0050}\\
        \bottomrule
    \end{tabular}
    \caption{Comparison of relative L2 errors under different normalization dimensions on the Airfoil, Darcy, and Elasticity datasets.}
    \label{tab:Softmax}
\end{table}

\subsection{Ablation Study}

\subsubsection{Effect of generalization and simplification steps.} 
We design ablation experiments to evaluate the impact of the generalization and simplification steps on model performance.
Specifically, we set $\psi^\top(\bm{K})=\text{Softmax}(\varphi^\top(\bm{Q}))$ to disable the generalization step, and we add a slice attention to disable the simplification step.
We compare these variants with LinearNO, as shown in Table~\ref{tab:Ablation}.
By comparing No.1 vs. No.2 and No.3 vs. No.4, we observe that the generalization step consistently improves model performance in most cases, confirming the benefit of using asymmetric projections.
Furthermore, the comparison between No.1 and No.3 shows that, under symmetric settings, slice attention does not consistently improve performance, which is also observed in Transolver.
In contrast, comparing No.2 and No.4 reveals that applying slice attention under asymmetric projections consistently hurts the performance.
Therefore, the simplification step is also justified.

\subsubsection{Slice Number.} The results in Table~\ref{tab:Slice_num} demonstrate that increasing the number of slices $M$ generally improves the prediction accuracy across different PDE tasks, with the most significant gains observed in tasks like NS and Elasticity. For most cases, the relative L2 error decreases as $M$ increases, with $M=256$ often yielding the best performance. However, the improvement exhibits diminishing returns beyond a certain point, and for some tasks such as Darcy and Airfoil, using an excessive number of slices might not lead to further improvement or may even degrade performance slightly. 
Therefore, selecting an optimal number of slices requires balancing accuracy gains with computational costs and task-specific characteristics.

\subsubsection{Softmax Normalization.}
We also conduct experiments to investigate the impact of normalization dimension on model performance.
Previous works present multiple combinations of normalization dimensions for $\varphi(\bm{Q})$ and $\psi(\bm{K})$~\cite{shen2021efficient,haoGNOTGeneralNeural2023}. In this section, compare different combinations of Softmax normalization. Here, we assume $\varphi(\bm{Q}), \psi(\bm{K}) \in \mathbb{R}^{N\times M}$.
As shown in Table~\ref{tab:Softmax}, applying Softmax to $\varphi(\bm{Q})$ and $\psi(\bm{K})$ on dimensions $M$ and $N$ performs best, because it maintains the row normalization property of the attention matrix $\varphi(\bm{Q})\psi^\top(\bm{K})$.

\section{Conclusion}
In this work, we first review the Physics-Attention and reveal its essence as a form of linear attention.
Based on this insight, we propose a two-step  transformation: a generalization step and a simplification step, which transform the original Physics-Attention into a more efficient model called LinearNO.
LinearNO has fewer parameters and lower computational cost. It achieves state-of-the-art results on all six
standard PDE benchmark tasks.
In addition, LinearNO demonstrates superior performance on two industrial-level datasets compared to other models.
We further show through experiments that our model achieves a higher rank utilization than Transolver.
Finally, we conduct a series of ablation studies to validate the effectiveness. %

\section{Acknowledgments}

This work was supported by National Key R\&D Program 2025YFB3003600 and the Open Fund of National Key Laboratory of Parallel and Distributed Computing (PDL)NO.WDZC20255290101.
\bibliography{aaai2026}

\clearpage
\section*{Appendix}
%
%
\makeatletter
\@ifundefined{isChecklistMainFile}{
  \newif\ifreproStandalone
  \reproStandalonetrue
}{
  \newif\ifreproStandalone
  \reproStandalonefalse
}
\makeatother
\reproStandalonefalse
\ifreproStandalone

\def\aaaianonymous{true}
%
%

\documentclass[letterpaper]{article} 

\ifdefined\aaaianonymous
    \usepackage[submission]{aaai2026}  
\else
    \usepackage{aaai2026}              
\fi

\usepackage{amsmath}
\usepackage{amsthm}
\newtheorem{theorem}{Theorem}
\usepackage{amsfonts}
\usepackage{subcaption}
\usepackage{booktabs}
\usepackage{multirow}
\usepackage{tabularx}
\usepackage{amssymb}
\usepackage{makecell}
\usepackage{bm}
\newcolumntype{Y}{>{\centering\arraybackslash}X}
\usepackage{times}  
\usepackage{helvet}  
\usepackage{courier}  
\usepackage[hyphens]{url}  
\usepackage{graphicx} 
\urlstyle{rm} 
\def\UrlFont{\rm}  
\usepackage{natbib}  
\usepackage{caption} 
\frenchspacing  
\setlength{\pdfpagewidth}{8.5in} 
\setlength{\pdfpageheight}{11in} 

\usepackage{algorithm}
\usepackage{algorithmic}

\usepackage{newfloat}
\usepackage{listings}
\DeclareCaptionStyle{ruled}{labelfont=normalfont,labelsep=colon,strut=off} 
\lstset{%
	basicstyle={\footnotesize\ttfamily},
	numbers=left,numberstyle=\footnotesize,xleftmargin=2em,
	aboveskip=0pt,belowskip=0pt,
	showstringspaces=false,tabsize=2,breaklines=true}
\floatstyle{ruled}
\newfloat{listing}{tb}{lst}{}
\floatname{listing}{Listing}

\pdfinfo{
/TemplateVersion (2026.1)
}

\setcounter{secnumdepth}{0} 

\ifdefined\aaaianonymous
    \title{Supplementary Material of\\Transolver is a Linear Transformer: Revisiting Physics-Attention through the Lens of Linear Attention}
\else
    \title{AAAI 2026 Supplementary Material\\Camera Ready}
\fi

\ifdefined\aaaianonymous
\author{
    Anonymous Submission
}
\affiliations{
}
\else
\author{
    Wenjie Hu\textsuperscript{\rm 1},
    Sidun Liu\textsuperscript{\rm 1},
    Qiao Peng\textsuperscript{\rm 1},
    Zhenglun Sun\textsuperscript{\rm 1},
    Yong Dou\textsuperscript{\rm 1}
}
\affiliations{
    \textsuperscript{\rm 1}National University of Defense Technology,
    Changsha,
    China\\


    
    \{hwjb127,liusidun,pengqiao,zhenglun\_sun,yongdou\}@nudt.edu.cn
%
}
\fi

\begin{document}
\fi
\maketitle


\section{A. Proof of Theorem 1}
In this section, we present the complete proof of Theorem 1, following the proof techniques proposed by \citeauthor{DBLP:conf/icml/ChenW24}~\shortcite{DBLP:conf/icml/ChenW24}.
\begin{proof}
    For any $x \in \Omega$, define
\begin{equation}
    \begin{aligned}
    \mathcal{G}_n(x):= \frac{1}{N}\sum_{k=1}^{N} \varphi(v({x})) \exp({B}^{\top}v({x_k})^{\top})(v(x_k)R)
    \end{aligned}
\end{equation}

\begin{equation}
    \begin{aligned}
    \mathcal{H}_n:= \frac{1}{N}\sum_{j=1}^{N}\exp{({B}^{\top}v({x_j})^{\top})}
    \end{aligned}
\end{equation}
    Then, we have
\begin{equation}
    \begin{aligned}
        \text{LinearNO}(x_i) &= \sum_{k=1}^{N} \frac{\varphi(v({x_i})) \exp({B}^{\top}v({x_k})^{\top})}{\sum_{j=1}^{N}\exp{({B}^{\top}v({x_j})^{\top})}} (v(x_k)R) \\
        &= \frac{\frac{1}{N}\sum_{k=1}^{N}\varphi(v({x_i})) \exp({B}^{\top}v({x_k})^{\top}(v(x_k)R))}{\frac{1}{N}\sum_{j=1}^{N}\exp{({B}^{\top}v({x_j})^{\top})}}  \\
        &= \frac{\mathcal{G}_n(x_i)}{\mathcal{H}_n}=:\mathcal{F}_n(x_i) \\
    \end{aligned}
\end{equation}
Note that $\mathcal{G}_n(x)$ and $\mathcal{H}_n$ are the Monte Carlo integration approximations to
\begin{equation}
    \begin{aligned}
        \mathcal{G}(x) &= \int_{\Omega} \varphi(v(x)) \exp({B}^{\top}v(y)^{\top})(v(y)R) d \mu_{\Omega}(y) \\
        \mathcal{H} &= \int_{\Omega} \exp({B}^{\top}v(y)^{\top}) d \mu_{\Omega}(y)
    \end{aligned}
\end{equation}
respectively. By the strong law of large numbers, we know that
\begin{equation}
    \begin{aligned}
        \Pr\left\{\lim _{n \rightarrow \infty} \mathcal{G}_n(x) = \mathcal{G}(x) \right\}=1 \\
        \Pr\left\{\lim _{n \rightarrow \infty} \mathcal{H}_n = \mathcal{H} \right\}=1
    \end{aligned}
\end{equation}
It follows that
\begin{equation}
    \begin{aligned}
    \mathcal{F}_n(x)-\mathcal{F}(x) &= \frac{\mathcal{G}_n(x)}{\mathcal{H}_n}-\frac{\mathcal{G}(x)}{\mathcal{H}} \xrightarrow{\text { a.s. }} 0\\
    \label{eq:monte}
    \end{aligned}
\end{equation}
    Define
\begin{equation}
    \begin{aligned}
        A_{n}:&=\int_{\Omega}\left|\frac{\mathcal{G}_{n}(x)}{\mathcal{H}_{n}}-\frac{\mathcal{G}(x)}{\mathcal{H}}\right| d \mu_{\Omega}(x) \\
        &=\int_{\Omega}\left|\mathcal{F}_{n}(x)-\mathcal{F}(x)\right| d \mu_{\Omega}(x)
    \end{aligned}
\end{equation}
Then, for any  $\varepsilon>0$ , we have
\begin{equation}
    \begin{aligned}
\operatorname{Pr}\left\{\lim _{n \rightarrow \infty} A_{n}=0\right\}=1, \quad \lim _{n \rightarrow \infty} \operatorname{Pr}\left\{A_{n} \leq \frac{\varepsilon}{2}\right\}=1
    \label{eq:limit}
    \end{aligned}
\end{equation}
Using the Chebychev inequality and the elementary inequality $ (p-q)^{2} \leq p^{2}+q^{2} $ for non-negative$  p $ and $ q $, we have
\begin{equation}
\begin{aligned}
&\operatorname{Pr}\left\{\left|\frac{1}{N} \sum_{i=1}^{N}\right| \mathcal{F}_{n}\left(x_{i}\right)-\mathcal{F}\left(x_{i}\right)\left|-A_{n}\right|>\frac{\varepsilon}{2}\right\} \\
& \leq \frac{4}{N \varepsilon^{2}} \int_{\Omega}\left(\left|\mathcal{F}_{n}(x)-\mathcal{F}(x)\right|-A_{n}\right)^{2} d \mu_{\Omega}(x) \\
& \leq \frac{4}{N \varepsilon^{2}}\left(A_{n}^{2}+\int_{\Omega}\left(\mathcal{F}_{n}(x)-\mathcal{F}(x)\right)^{2} d \mu_{\Omega}(x)\right)
\end{aligned}
\end{equation}
which further yields
\begin{equation}
\begin{aligned}
&\operatorname{Pr}\left\{\left|\frac{1}{N} \sum_{i=1}^{N}\right| \mathcal{F}_{n}\left(x_{i}\right)-\mathcal{F}\left(x_{i}\right)\left|-A_{n}\right| \leq \frac{\varepsilon}{2}\right\} \\
&\geq 1-\frac{4}{N \varepsilon^{2}}\left(A_{n}^{2}+\int_{\Omega}\left(\mathcal{F}_{n}(x)-\mathcal{F}(x)\right)^{2} d \mu_{\Omega}(x)\right) .
\end{aligned}
\end{equation}
Because
\begin{equation}
\begin{aligned}
&\operatorname{Pr}\left\{\frac{1}{N} \sum_{i=1}^{N}\left|\mathcal{F}_{n}\left(x_{i}\right)-\mathcal{F}\left(x_{i}\right)\right|
\leq A_{n}+\frac{\varepsilon}{2}\right\} \\
&\geq \operatorname{Pr}\left\{\left|\frac{1}{N} \sum_{i=1}^{N}\right| \mathcal{F}_{n}\left(x_{i}\right)-\mathcal{F}\left(x_{i}\right)\left|-A_{n}\right| \leq \frac{\varepsilon}{2}\right\},
\end{aligned}
\end{equation}
we obtain
\begin{equation}
\begin{aligned}
&\operatorname{Pr}\left\{\frac{1}{N} \sum_{i=1}^{N}\left|\mathcal{F}_{n}\left(x_{i}\right)-\mathcal{F}\left(x_{i}\right)\right| \leq A_{n}+\frac{\varepsilon}{2}\right\} \\
&\geq 1-\frac{4}{N \varepsilon^{2}}\left(A_{n}^{2}+\int_{\Omega}\left(\mathcal{F}_{n}(x)-\mathcal{F}(x)\right)^{2} d \mu_{\Omega}(x)\right)
\label{eq:21}
\end{aligned}
\end{equation}
Notice that if
$$\frac{1}{N} \sum_{i=1}^{N}\left|\mathcal{F}_{n}\left(x_{i}\right)-\mathcal{F}\left(x_{i}\right)\right| \leq A_{n}+\frac{\varepsilon}{2} \quad \text { and } \quad A_{n} \leq \frac{\varepsilon}{2},
$$
then
$$\frac{1}{N} \sum_{i=1}^{N}\left|\mathcal{F}_{n}\left(x_{i}\right)-\mathcal{F}\left(x_{i}\right)\right| \leq \varepsilon$$
This implies
\begin{equation}
\begin{aligned}
&\operatorname{Pr}\left\{\frac{1}{N} \sum_{i=1}^{N}\left|\mathcal{F}_{n}\left(x_{i}\right)-\mathcal{F}\left(x_{i}\right)\right| \leq \varepsilon\right\} \\
& \geq \operatorname{Pr}\left\{\frac{1}{N} \sum_{i=1}^{N}\left|\mathcal{F}_{n}\left(x_{i}\right)-\mathcal{F}\left(x_{i}\right)\right| \leq A_{n}+\frac{\varepsilon}{2} \text { and } A_{n} \leq \frac{\varepsilon}{2}\right\} \\
& \geq \operatorname{Pr}\left\{\frac{1}{N} \sum_{i=1}^{N}\left|\mathcal{F}_{n}\left(x_{i}\right)-\mathcal{F}\left(x_{i}\right)\right| \leq A_{n}+\frac{\varepsilon}{2}\right\}\\
&+\operatorname{Pr}\left\{A_{n} \leq \frac{\varepsilon}{2}\right\}-1
\end{aligned}
\end{equation}
where the second step follows from the probability inequality  $\operatorname{Pr}(a \cap b) \geq \operatorname{Pr}(a)+\operatorname{Pr}(b)-1$ . Combing it with Eq.~\eqref{eq:21}, we obtain

\begin{equation}
\begin{aligned}
&\operatorname{Pr}\left\{\frac{1}{N} \sum_{i=1}^{N}\left|\mathcal{F}_{n}\left(x_{i}\right)-\mathcal{F}\left(x_{i}\right)\right| \leq \varepsilon\right\} \\
&\geq \operatorname{Pr}\left\{A_{n} \leq \frac{\varepsilon}{2}\right\}\\
&-\frac{4}{N \varepsilon^{2}}\left(A_{n}^{2}+\int_{\Omega}\left(\mathcal{F}_{n}(x)-\mathcal{F}(x)\right)^{2} d \mu_{\Omega}(x)\right)
\end{aligned}
\end{equation}
Taking  $n \rightarrow+\infty$  and using Eq.~\eqref{eq:monte}-\eqref{eq:limit}, we obtain
$$1 \geq \lim _{n \rightarrow+\infty} \operatorname{Pr}\left\{\frac{1}{N} \sum_{i=1}^{N}\left|\mathcal{F}_{n}\left(x_{i}\right)-\mathcal{F}\left(x_{i}\right)\right| \leq \varepsilon\right\} \geq 1$$
Therefore,
$$\lim _{n \rightarrow+\infty} \operatorname{Pr}\left\{\frac{1}{N}\left\|\operatorname{LinearNO}\left(x\right)-\mathcal{F}(x)\right\| \leq \varepsilon\right\}=1$$

for any  $\varepsilon>0$ . This means the LinearNO converges in probability to the continuous integral operator. The proof is completed.

\end{proof}

\section{B. Dataset and Implementation Details}
In this section, we provide a detailed description of our datasets and experimental setup.
\subsection{Dataset details}
Table~\ref{tab:Dataset} summarizes the key information of the datasets. Detailed descriptions are provided below.
\begin{table}[t]
    \centering
    \begin{tabular}{l|c|c}
        \toprule
        Dataset &Grid Type &Grid Size  \\
        \midrule
        Airfoil~\shortcite{DBLP:journals/jmlr/LiHLA23} &Structure Grid&11,271\\
        Pipe~\shortcite{DBLP:journals/jmlr/LiHLA23} &Structure Grid &16,641\\
        Plasticity~\shortcite{DBLP:journals/jmlr/LiHLA23} &Structure Grid &3,131\\
        \midrule
        NS~\shortcite{liFourierNeuralOperator2021} &Cartesian Grid &4,096\\
        Darcy~\shortcite{liFourierNeuralOperator2021} &Cartesian Grid &7,225\\
        Burgers~\shortcite{wangLatentNeuralOperator2024} &Cartesian Grid &16,384\\
        \midrule
        Elasticity~\shortcite{DBLP:journals/jmlr/LiHLA23} &Unstructure Grid &972\\  
        AirfRANS~\shortcite{bonnet2022airfrans} &Unstructure Grid &32,000\\
        Shape-Net Car~\shortcite{DBLP:journals/tog/ShapeCar} &Unstructure Grid&32,186\\
        \bottomrule
    \end{tabular}
    \caption{Datasets summary.}
    \label{tab:Dataset}
\end{table}

\noindent\textbf{Airfoil}~\cite{DBLP:journals/jmlr/LiHLA23}
In this benchmark, the input airfoil geometries are discretized onto a structured grid of size $221 \times 51$. The input is the coordinates of each grid point ($221 \times 51 \times 2$) and the output is the Mach number at each grid point ($221 \times 51 \times 1$). All geometries are generated by deforming the baseline airfoil NACA-0012. We use 1,000 samples for training and 200 samples for testing.

\noindent\textbf{Pipe}~\cite{DBLP:journals/jmlr/LiHLA23}
In this benchmark, the pipe geometry is discretized onto a structured grid of size $129 \times 129$. The input consists of the coordinates of each grid point ($129 \times 129 \times 2$), and the output is the velocity at each point ($129 \times 129 \times 1$). The pipe geometries are generated by controlling the centerline of the pipe. A total of 1,000 samples are used for training and 200 samples for testing.

\noindent\textbf{Darcy}~\cite{liFourierNeuralOperator2021}
In this benchmark, the process is discretized into $421  \times  421$ Cartesian grid and then downsampled to a resolution of $85  \times  85$ for the main experiments. The input consists of the coordinates and the porous medium of each point ($85  \times  85  \times  (2+1)$), while the output is the pressure at each point ($85  \times  85  \times  1$). For training, 1000 samples are used, 200 samples are generated for testing, and different cases contain different medium structures.

\noindent\textbf{Navier-Stokes (NS)}~\cite{liFourierNeuralOperator2021}
In this benchmark, the flow field is discretized on a 
$64 \times 64$ Cartesian grid. The input consists of the velocity fields from the past 10 frames, with a shape of 
$64 \times 64 \times 10 \times 1$, and the output is the velocity fields for the next 10 frames with a shape of 
$64 \times 64 \times 10  \times 1$. The model predicts one frame at once and recursively generates the full 10-frame sequence in an autoregressive manner. For training, 1000 samples with different initial conditions are used, and 200 additional samples are generated for testing.

\noindent\textbf{Plasticity}~\cite{DBLP:journals/jmlr/LiHLA23}
This benchmark aims to model the future deformation behavior of a plastic material when compressed by a die of arbitrary shape.
For each sample, the arbitrarily-shaped die is discretized onto a structured grid of size 
$101 \times 31$. The input consists of the coordinates of each grid point and the corresponding external force, resulting in a tensor of shape 
$101 \times 31 \times (2+1)$. The output is the deformation of each point along four directions over the next 20 time steps, resulting in a tensor of shape $101 \times 31 \times 20  \times 4$. For training, 900 samples with various die shapes are used, and 80 additional samples are generated for testing.

\noindent\textbf{Elasticity}~\cite{DBLP:journals/jmlr/LiHLA23}
This benchmark focuses on predicting the internal stress distribution within an elastic material, given its structural configuration discretized into 972 points.
Each input sample is represented as a tensor of size $972  \times  2$, where each row corresponds to the 2D coordinates of a discretized point. The output provides the stress value at each of these locations, formatted as a tensor of shape $972  \times  1$. For training, 1000 structurally varied samples are utilized, with an additional 200 samples allocated for testing.
\begin{table*}[t]
    \centering
    \resizebox{\linewidth}{!}{
    \begin{tabular}{l|cccccccccc}
        \toprule
        Benchmarks &Loss & Epochs &LR &Optimizer &Scheduler &Batch &Layer &Heads &Dim &Slices\\
        \midrule
        Airfoil &rL2&500&0.001&AdamW&OneCycle &4&8&8&128&64 \\
        Pipe &rL2&500&0.001&AdamW&OneCycle &4&8&8&128&64 \\
        Plasticity&rL2&500&0.001&AdamW&OneCycle &8&8&8&128&64\\
        NS&rL2&500&0.001&AdamW&OneCycle &2&8&8&256&32\\
        Darcy&$\mathcal{L}_{rL2}+0.1\mathcal{L}_g$&500&0.001&AdamW&OneCycle &4&8&8&128&64\\
        Elasticity&rL2&500&0.001&AdamW&CosineAnnealing &1&8&8&128&64\\
        \midrule
        AirfRANS&$\mathcal{L}_{v}+0.5\mathcal{L}_s$&400&0.001&Adam&OneCycle &1&8&8&256&32\\
        Shape-Net Car&$\mathcal{L}_{v}+\mathcal{L}_s$&200&0.001&Adam&OneCycle &1&8&8&256&32\\
        \midrule
        Burgers&MSE&500&0.001&Adam&OneCycle &4&8&8&96&256\\
        \bottomrule
    \end{tabular}
    }
    \caption{Hyperparameters for different PDE solving tasks. $\mathcal{L}_{v}$ and $\mathcal{L}_s$ represent the relative L2 errors of the physical field in the surrounding region and on the surface, respectively.
$\mathcal{L}_g$ is the spatial gradient regularization term, and MSE denotes the mean squared error.}
    \label{tab:Impel}
\end{table*}

\begin{table*}[t]
    \centering
    
    \begin{tabularx}{\linewidth}{c|c|YYYYYY}
        \toprule
        Metric&Model & Airfoil &Pipe  &Plasticity & NS & Darcy& Elasticity \\
        \midrule
        \multirow{2}{*}{\makecell{Parameter\\(GB)}}     &Transolver &2.81&3.07&2.84&11.23&2.83&0.71\\
                                                        &LinearNO &\textbf{1.77}&\textbf{1.77}&\textbf{1.80}&\textbf{3.38}&\textbf{1.77}&\textbf{0.59}\\
        \midrule
        \multirow{2}{*}{\makecell{Computer Cost\\(GFLOPs)}}     &Transolver &32.38&52.17&9.10&46.16&20.87&0.76\\
                                                                &LinearNO   &\textbf{21.34}&\textbf{31.51}&\textbf{6.03}&\textbf{15.53}&\textbf{13.68}&\textbf{0.69}\\
        \bottomrule
    \end{tabularx}
    \caption{Full comparison on parameter count and computational cost between LinearNO and Transolver.}
    \label{tab:ap_Param}
\end{table*}
\noindent\textbf{AirfRANS}~\cite{bonnet2022airfrans}
It's a dataset for studying the two-dimensional incompressible steady-state Reynolds-Averaged Navier-Stokes equations over airfoils at a subsonic regime and for different angles of attack.
Unlike the Airfoil dataset~\cite{DBLP:journals/jmlr/LiHLA23}, this dataset features more complex geometries, a larger number of grid points, richer input variables, and requires predicting more variables. Specifically, it adopts airfoil shapes from the 4 and 5-digit series defined by the National Advisory Committee for Aeronautics (NACA).
Each sample consists of approximately 32,000 unstructured grid points covering both the airfoil surface and the surrounding flow region, associated with various flow conditions such as different angles of attack and Reynolds numbers. The input for each point includes the spatial coordinates, the inlet velocity, the Euclidean distance from the point to the airfoil, and the unit outward-pointing surface normal (set to zero for non-surface points), resulting in an input tensor of shape 
$32000 \times (2+2+1+2)$. The output includes the velocity, pressure, and turbulent kinematic viscosity at each point, with a shape of $32000 \times (2+1+1)$.
The lift coefficient ($C_L$) and the Spearman's correlation coefficient ($\rho_L$) can be derived from the predicted outputs. However, following the experimental setup proposed by Transolver~\cite{wu2024transolver}, we mainly focus on evaluating the pressure field in the surrounding volume (Volume), the surface pressure field on the airfoil (Surface), as well as the lift coefficient ($C_L$) and the Spearman's correlation coefficient ($\rho_L$).
By varying airfoil geometries, Reynolds numbers, and angles of attack, the AirfRANS dataset provides 1,000 samples in total, with 720 used for training, 80 for validation and 200 for testing.

\noindent\textbf{Shape-Net Car}~\cite{DBLP:journals/tog/ShapeCar}
This dataset simulates the physical fields around and on the surface of 889 different car shapes, each traveling at a speed of 72 km/h. Specifically, the car surface and its surrounding domain are discretized into 32,186 unstructured grid points. The input for each point includes the point position, signed distance function, and normal vector, resulting in an input shape of $32,186  \times  (3+1+3)$. The output consists of the velocity and pressure at each point, with a shape of $32,186  \times  (3+1)$. Similar to AirfRANS~\cite{bonnet2022airfrans}, we compute the drag coefficient ($C_D$) and the Spearman's correlation coefficient ($\rho_D$) based on the predicted physical fields. Following Transolver~\cite{wu2024transolver}, we focus on assessing the velocity field in the surrounding region (Volume), the surface pressure field (Surface), the drag coefficient ($C_D$), and the Spearman's correlation coefficient ($\rho_D$). We use 789 samples for training and 100 samples for testing.

\noindent\textbf{Burgers}~\cite{wangLatentNeuralOperator2024}: This dataset is used for the super-resolution experiments presented in Appendix E. Specifically, we adopt the 1D Burgers' equation as

\begin{equation}
    \begin{aligned}
\frac{\partial}{\partial t} u(x, t)=&0.01 \frac{\partial^{2}}{\partial x^{2}} u(x, t)-u(x, t) \frac{\partial}{\partial x} u(x, t), \\
u(x, 0) \sim& GP\left(0,\exp \left(-\frac{2}{p l^{2}} \sin ^{2}\left(\pi\left\|x-x^{\prime}\right\|\right)\right)\right), \\
u(0, t)=&u(1, t), x \in[0,1], t \in[0,1] \\
    \end{aligned}
\end{equation}
where $GP$ means Gaussian Process. Then, both time $t \in [0, 1]$ and space $x \in [0, 1]$ are discretized into a $128  \times  128$ grid with periodic boundary conditions. The initial conditions are sampled from a Gaussian process with periodic length $p = 1$ and scaling factor $l = 1$. The task is to randomly sample a certain number of points $(x', u(x'))$ from the domain $(x, t) \in [0, 1]  \times  [0.25, 0.75]$ as input, and reconstruct the solution $u(x)$ at the domain $(x, t) \in [0, 1]  \times  [0.25, 0.75]$. This setup is referred to as the Completer task in Section 4.2 of the LNO~\cite{wangLatentNeuralOperator2024}. We use 4096 samples for training, 128 samples for validating and 128 samples for testing.

\subsection{Metric}
\noindent\textbf{Relative L2 Error.} It is used as the evaluation metric for all benchmarks, which is defined as follows:
$$\text{Relative L2 Eorr}=\frac{\|y-\hat{y}\|_2}{\|y\|_2}$$
Where 
$y$ denotes the ground truth, $\hat{y}$ is the predicted value, and $\|\cdot\|_2$ denotes the L2 norm.

\noindent\textbf{Lift and Drag Coefficient ($C_L,C_D$).}  They are used as the evaluation metrics for AirfRANS and Shape-Net Car, which are defined as follows:
\begin{equation}
\mathbf{F} = \int_S \left( -p \mathbf{n} + \boldsymbol{\tau} \cdot \mathbf{n} \right) \, dS
\end{equation}

\begin{equation}
\boldsymbol{\tau} = \mu \left[ \nabla \mathbf{u} + (\nabla \mathbf{u})^\top \right]
\end{equation}
where $p$ is the pressure on the surface, $\mathbf{n}$ the unit normal vector pointing outward from the surface, $\boldsymbol{\tau}$ is the viscous stress tensor, $\mu$ is the dynamic viscosity of the fluid, and  $\mathbf{u}$ is the velocity field,. The total force $\mathbf{F}$ is projected onto the streamwise and vertical directions to obtain drag $F_D$ and lift $F_L$, respectively:
\begin{equation}
F_D = \mathbf{F} \cdot \mathbf{e}_x, \quad
F_L = \mathbf{F} \cdot \mathbf{e}_y
\end{equation}
The dimensionless lift and drag coefficients are defined as:
\begin{equation}
C_L = \frac{F_L}{\frac{1}{2} \rho U_\infty^2 A}, \quad
C_D = \frac{F_D}{\frac{1}{2} \rho U_\infty^2 A}
\end{equation}
where the fluid density is denoted by $\rho$, $U_\infty^2s$ is the freestream velocity magnitude, and 
$A$ is the reference area of the body (e.g., chord length in 2D or projected frontal area in 3D). The lift and drag coefficients accurately reflect the aerodynamic performance of the airfoil and car, respectively.

\noindent\textbf{Spearman's Correlation Coefficient ($\rho$).} Given $n$ samples, the predicted force coefficients are denoted as $\hat{C}=\{\hat{C}^1,\dots,\hat{C}^n\}$ , and the ground truth are denoted as $C=\{C^1,\dots,C^n\}$. The Spearman's correlation coefficient $\rho$ is defined as follows:
\begin{equation}
\rho=\frac{\operatorname{Cov}(\operatorname{rank}(C), \operatorname{rank}(\hat{C}))}{\sigma_{\operatorname{rank}(C)} \sigma_{\operatorname{rank}(\hat{C})}}
\end{equation}
where $\operatorname{Cov}(\cdot,\cdot)$ is the covariance function, $\operatorname{rank}(\cdot)$ is the ranking function, and $\sigma_{\operatorname{rank}(\cdot)}$ is the standard deviation of these ranks. 
The Spearman's correlation coefficient $\rho$ is used to quantify the monotonic relationship between the true and predicted force coefficients, where a value closer to 1 reflects stronger predictive consistency.

\subsection{Implementation Details}\label{Impeldetails}
The detailed hyperparameters are listed in Table~\ref{tab:Impel}. To ensure a fair comparison, we follow the experimental settings of Transolver~\cite{wu2024transolver} for the six standard PDE benchmarks, AirfRANS, and Shape-Net Car. For the Burgers dataset, we adopt the settings from LNO~\cite{wangLatentNeuralOperator2024}. All models are trained three times on a NVIDIA RTX 4090 GPU. Following Transolver, 
we adopt a single convolutional layer as a preprocessing step for datasets defined on structured and Cartesian grids, except for the NS dataset, as it corresponds to an unsteady fluid flow problem.

\section{C. Fully Experimental Results}
In this section, we will present the experimental results that were not fully displayed in the main paper. 

\subsection{Efficiency Analysis}
The detailed comparison of the total number of parameters and computational complexity with batch size of 1 between LinearNO and Transolver on six standard datasets is summarized in Table~\ref{tab:ap_Param}. While Transolver has a complexity of $\mathcal{O}(NM + M^2)$, our simplification reduces the complexity to $\mathcal{O}(NM)$, where $N$ is the number of input points and $M$ is the number of slices.

\subsection{Effect of generalization and simplification steps}
Here, we provide the results of all 6 datasets for the ablation study on the generalization step and simplification step, as shown in Table~\ref{tab:ap_Ablation}. The conclusion is consistent with the description in the main paper, proving the effectiveness of generalization and simplification steps.

\begin{table*}
    \centering
    \begin{tabularx}{\linewidth}{c|cc|YYYYYY}
        \toprule
       No.&\textit{Gen.}  & \textit{Sim.} & Airfoil &Pipe  &Plasticity & NS & Darcy& Elasticity \\
        \midrule
        1&$ \times $ & $ \times $ & 0.0067&0.0029&0.0016  &0.1261& 0.0064& 0.0069\\
        2&$\checkmark$ & $ \times $ & 0.0054&0.0032 &0.0013     &0.0711   & 0.0061& 0.0062\\
        3&$ \times $ & $\checkmark$ & 0.0071&0.0030 &0.0015     &0.1229 & 0.0052& 0.0064\\
        4&$\checkmark$ & $\checkmark$ & \textbf{0.0049} &\textbf{0.0024}&\textbf{0.0011}&\textbf{0.0699}& \textbf{0.0050}& \textbf{0.0050}\\
        \bottomrule
    \end{tabularx}
    \caption{Full ablation study of generalization and simplification. \textit{Gen.} denotes the generalization step, and \textit{Sim.} denotes the simplification step.}
    \label{tab:ap_Ablation}
\end{table*}

\begin{table*}
    \centering
    \begin{tabularx}{\linewidth}{c|YYYYYY}
        \toprule
        Num of Slice $M$ & Airfoil &Pipe  &Plasticity & NS & Darcy& Elasticity\\
        \midrule
        16  &0.0052&0.0030&0.0013& 0.0771&0.0056   &0.0062\\
        32  &0.0053&0.0028&0.0014&0.0699&0.0053&0.0056\\
        48  &0.0054&0.0029&0.0012&0.0691&0.0051&0.0052\\
        64  &0.0049&0.0024&0.0011&0.0650&0.0050&0.0050\\
        96  &0.0052&0.0026&0.0011&0.0643&\textbf{0.0048}&0.0056\\
        128 &\textbf{0.0047}&0.0026&0.0011&0.0612&0.0050&0.0051\\
        256&0.0050&\textbf{0.0022}&\textbf{0.0010}&\textbf{0.0563}&0.0049&\textbf{0.0046}\\
        \bottomrule
    \end{tabularx}
    \caption{Relative L2 errors of LinearNO with different numbers of slices $M$.}
    \label{tab:ap_Slice_num}
\end{table*}
\begin{table*}
    \centering
    \begin{tabular}{l|cc|cc}
        \toprule
        \multirow{2}{*}{Dataset}   &\multicolumn{2}{c}{OOD REYNOLDS} &\multicolumn{2}{c}{OOD ANGLES}\\ 
        \cmidrule{2-3} \cmidrule{4-5}
        & Reynolds Ranges &Samples   & AoA Range      &Samples       \\
        \midrule
        Training Set & $[3  \times  10^6, 5 \times 10^6]$ & 500 & $[-2.5^{\circ}, 12.5^{\circ}]$ & 800\\
            Test Set & \mbox{$[2  \times  10^6, 3 \times 10^6]\cup[5 \times 10^6,6 \times 10^6]$} & 200 & $[-5^{\circ}, 2.5^{\circ}]\cup[12.5^{\circ},15^{\circ}]$ & 200\\
        \bottomrule
    \end{tabular}
    \caption{Experimental configuration for OOD generalization.}
    \label{tab:OODdataset}
\end{table*}
\begin{table*}
    
    \centering
    \begin{tabularx}{\linewidth}{l|YYYYYYYY}
        \toprule
        \multirow{2}{*}{Model}   &\multicolumn{4}{c}{OOD REYNOLDS} &\multicolumn{4}{c}{OOD ANGLES}\\ 
        \cmidrule{2-5} \cmidrule{6-9}
        & Volume $\downarrow$   & Surface $\downarrow$      & $C_L \downarrow$         & $\rho_L \uparrow$  & Volume $\downarrow$   & Surface $ \downarrow$  & $C_L \downarrow$  & $\rho_L \uparrow$\\
        \midrule
        MLP        &0.0669 &0.1153 &0.6205 &0.9578 &0.1309 &0.3311 &0.4128 &0.9572\\
        GraphSage \shortcite{DBLP:conf/nips/HamiltonYL17}  &0.0798 &0.1254 &0.4333 &0.9707 &0.1192 &0.2359 &0.2538 &0.9894\\
        PointNet \shortcite{DBLP:conf/cvpr/QiSMG17}&0.0838 &0.1403 &0.3836 &0.9806 &0.2021 &0.4649 &0.4425 &0.9784\\
        GraphUNet \shortcite{DBLP:conf/icml/GaoJ19}&0.0538 &0.1168 &0.4664 &0.9645 &0.0979 &0.2391 &0.3756 &0.9816\\
        MeshGraphNet \shortcite{DBLP:conf/iclr/PfaffFSB21}&0.2789 &0.2382 &1.7718 &0.7631 &0.4902 &1.1071 &0.6525 &0.8927\\
        GNO \shortcite{liNeuralOperatorGraph2020}         &0.0833 &0.1562 &0.4408 &0.9878 &0.1626 &0.2359 &0.3038 &0.9884\\
        
        \midrule
        GALERKIN \shortcite{DBLP:conf/nips/Cao21}    &0.0330 &0.0972 &0.4615 &0.9826 &0.0577 &0.2773 &0.3814 &0.9821\\
        GNOT \shortcite{haoGNOTGeneralNeural2023}        &0.0305 &0.0959 &0.3268 &0.9865 &0.0471 &0.3466 &0.3497 &0.9868\\
        GINO \shortcite{DBLP:conf/nips/LiKCLKONS0AA23}        &0.0839 &0.1825 &0.4180 &0.9645 &0.1589 &0.2469 &0.2583 &0.9923\\    
        LNO$^*$ \shortcite{wangLatentNeuralOperator2024}        & 0.0825     &0.1762         &0.7440&0.9399     &\textbf{0.0346}  & \textbf{0.0790} & 0.3732 &0.9905\\
        Transolver$^*$ \shortcite{wu2024transolver} &\underline{0.0122} & \underline{0.0550}& \textbf{0.1622}& \underline{0.9904}  &0.0480 &  \underline{0.2335} & \underline{0.2438} & \underline{0.9948}\\
        \midrule
        LinearNO (ours)   &\textbf{0.0112} & \textbf{0.0372}  & \underline{0.2400}& \textbf{0.9951}&\underline{0.0464} & {0.2500}& \textbf{0.0987}& \textbf{0.9963}\\
        
        \bottomrule
    \end{tabularx}
    \caption{Performance OOD Generalization on AirfRANS Dataset.
    Volume and Surface represent the physical fields in the surrounding flow region and on the object surface, respectively.
$C_L$ and $C_D$ denote the lift coefficient and drag coefficient, respectively.
The table reports their relative L2 errors.
Spearman's correlation coefficient $\rho$ is closer to 1, indicating better performance.
$^*$ indicates results reproduced by us.
An underscore indicates the second-best result, and bold indicates the best result.}
    \label{tab:OOD}
\end{table*}

\subsection{Ablation on the Number of Slices}
The full results on all 6 datasets, where the slice number varies from 16 to 256, are listed in Table~\ref{tab:ap_Slice_num}. Different PDE solving tasks exhibit different optimal numbers of slices, indicating varying intrinsic complexities across problems. Overall, using more slices generally leads to higher accuracy, but also incurs greater computational cost. This suggests that the number of slices should be selected appropriately based on the specific problem.

\section{D. OOD Generalization on AirfRANS Dataset}
In this section, generalization experiments are conducted on the AirfRANS dataset. Specifically, the training and testing sets have different ranges of Reynolds numbers and angles of attack, and this experimental configuration is presented in Table~\ref{tab:OODdataset}. The overall experimental results are shown in Table~\ref{tab:OOD}.
In both OOD scenarios, the model ranks first in 6 metrics and second in 2 metrics. It demonstrates particularly significant advantages in terms of the $\rho_L$ correlation (reflecting the consistency of prediction trends) and the $C_L$ error (accuracy of aerodynamic coefficients), indicating that it has the strongest adaptability to out-of-distribution conditions.

\section{E. Super-Resolution Experiment}

\begin{figure}
    \centering
    \includegraphics[width=0.6\linewidth]{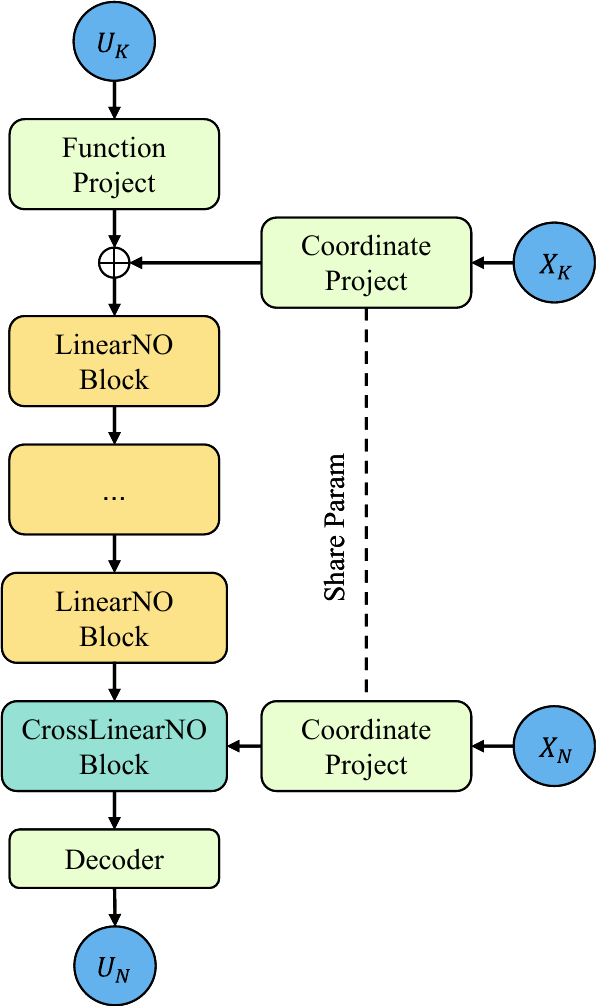}
    \caption{Overall Architecture with Cross-LinearNO.}
    \label{fig:crossLinearNO}
\end{figure}

To evaluate the discretization-invariance of LinearNO, we conduct a super-resolution experiment using the 1D Burgers dataset introduced in Appendix B. This task follows a sequence-to-sequence (seq2seq) setting, where the input coordinates and query coordinates are different. To adapt to this task, we extend the overall architecture as illustrated in Figure~\ref{fig:crossLinearNO}. Specifically, we decouple the encoder and employ Coordinate Project and Function Project for the input coordinates $\bm{X}_K \in \mathbb{R}^{K \times 2}$ and solution function $\bm{U}_K \in \mathbb{R}^{K \times 1}$ respectively, and their sum yields $\bm{H}_K \in \mathbb{R}^{K \times d_h} $.The same coordinate projection is then used on query  coordinates $\bm{X}_N \in \mathbb{R}^{N \times 2}$ to obtain $\bm{H}_N \in \mathbb{R}^{N \times d_h}$. Note that $\bm{X}_K$ is randomly sampled from $\bm{X}_N$,  and therefore $\bm{X}_K \subseteq \bm{X}_N$. We further propose a Cross-LinearNO Block, a new module capable of accepting the projected query coordinates and computing cross-linear attention in the functional space, which can be formalized as follows:
\begin{equation}
    \begin{aligned}
        \varphi(\bm{Q}) =& \text{Softmax}(\text{Linear}^{(Q)}(\bm{H}_N)) \\
        \psi^\top(\bm{K}) =& \text{Softmax}(\text{Linear}^{(K)}(\bm{H}^{(L)}_K)^\top)\\
        \text{Cross-LinearNO}&(\bm{H}_N, \bm{H}^{(L)}_K) = \\
        &\varphi(\bm{Q})\left(\psi^\top(\bm{K})\cdot \text{Linear}^{(V)}(\bm{H}_K)\right)
        \label{eq:phi_sp}
    \end{aligned}
\end{equation}

Here, $\bm{H}^{(L)}_K$ represents the hidden representation obtained after $L$ layers of the LinearNO Block. We selected three baseline models—DeepONet~\cite{luLearningNonlinearOperators2021}, GNOT~\cite{haoGNOTGeneralNeural2023}, and LNO~\cite{wangLatentNeuralOperator2024}—to compare the reconstructed physical field against the original one using the Relative MAE metric under various sampling rates: 20\%, 10\%, 5\%, 1\%, and 0.5\%. Since Transolver does not perform the decoupling operation, it cannot be applied to this task. The baseline settings follow those described in Appendix D of LNO~\cite{wangLatentNeuralOperator2024}. To ensure a fair comparison, our model adopts the same hidden dimension and number of layers as LNO, as shown in Table~\ref{tab:Impel}. As shown in the Table~\ref{tab:Completer}, our model achieves the best performance for four out of five sampling rates, demonstrating the competitiveness of LinearNO on the super-resolution task.

\begin{table}
    \centering
    \resizebox{\linewidth}{!}{
    \begin{tabular}{l|ccccc}
        \toprule
        \multirow{2}{*}{Model}   &\multicolumn{5}{c}{Sampling rates} \\ 
        \cmidrule{2-6} 
        &20\% &10\%  &5\% & 1\% & 0.5\%\\
        \midrule
        DeepONet &2.51\% &2.59\% &2.82\% &3.25\% &4.82\%\\
        GNOT &1.12\% &1.39\% &1.62\% &1.63\% &2.56\%\\
        LNO$^*$ &0.75\% &\textbf{0.48\%} &1.21\% &1.91\% &2.10\%\\
        \midrule
        LinearNO&\textbf{0.73\%} &0.71\% &\textbf{0.78\%} &\textbf{1.24\%} &\textbf{1.51\%}\\
        \bottomrule
    \end{tabular}
    }
    \caption{Relative MAE of Model with different settings of sampling rates. $^*$ indicates results reproduced by us.}
    \label{tab:Completer}
\end{table}

\section{F. Equivalence of Cross-Attention with Learnable Tokens to Linear Attention}

Cross-attention is originally introduced to model interactions between two sequences, allowing one sequence to attend to relevant information in the other. However, in cross-attention with learnable tokens, most existing works treat these tokens as coordinate anchors in a latent space, which are used to compress the original sequence \cite{wangLatentNeuralOperator2024,serranoAROMAPreservingSpatial2024,alkinUniversalPhysicsTransformers2024}. Taking the Physics-Cross-Attention module in LNO \cite{wangLatentNeuralOperator2024} as an example, the operation can be formulated as follows:
\begin{equation}
    \begin{aligned}
        \bm{S}_M&=\text{Physics-Cross-Attention}(\bm{Z}_M,\bm{H}_N,\bm{\hat{H}}_N)\\
        &=\text{Softmax}\left(\frac{1}{\sqrt{d_h}} \bm{Z_M} {{W}_{q}} {{W}_{k}}^{\top} \bm{H}_N^{\top}\right) \bm{\hat{H}}_N {W}_{v}\\
        &=\text{Softmax}\left({W}_{1} \bm{H}_N^{\top}\right) \bm{\hat{H}}_N {W}_{v}\\
        &=\text{Softmax}((\text{Linear}^{(K)}(\bm{H}_N)^{\top})\text{Linear}^{(V)}(\bm{\hat{H}}_N)   
        \label{eq:cross_encode}
    \end{aligned}
\end{equation}
\begin{equation}
    \begin{aligned}
         \bm{S}_M'= \text{Self-Attention}\circ \dots \circ \text{Self-Attention}(\bm{S}_M)
    \end{aligned}
\end{equation}
\begin{equation}
    \begin{aligned}
        \bm{H}_N'&=\text{Physics-Cross-Attention}(\bm{H}_N,\bm{Z}_M,\bm{S}'_M)\\
        &=\text{Softmax}\left(\frac{1}{\sqrt{d_h}} \bm{H}_N {W}_{q}^{\prime} {W}_{k}^{\top} \bm{Z}_M^{\top}\right) \bm{S}'_M {W}_{v}^{\prime}\\
        &=\text{Softmax}\left(\bm{H}_N {W}_{2}^{\top}\right) \bm{S}'_M {W}_{v}^{\prime} \\
        &=\text{Softmax}(\text{Linear}^{(Q)}(\bm{H}_N))\bm{S}'_M{W}_{v}^{\prime} 
        \label{eq:cross_decode}
    \end{aligned}
\end{equation}
where $\bm{Z}_M=\{\bm{z}_i\}_{i=1}^M \in \mathbb{R}^{M\times d_h}$ denotes the learnable tokens. $W_q, W_k, W_v,W'_q, W'_k,$ and $ W'_v$ are learnable matrices in $\mathbb{R}^{d_h\times d_h}$. By the associativity of matrix multiplication, the matrices $\frac{1}{\sqrt{d_h}}{Z_M} {W}_{q} {W}_{k}^{\top}$ and $\frac{1}{\sqrt{d_h}}{W}_{q}^{\prime} {W}_{k}^{\top} {Z_M}^{\top}$ can be merged into two learnable matrices, denoted as $W_1$ and $W_2$.
In LNO, $\bm{H}_N$ contains only coordinate information, while $\bm{\hat{H}}_N$ includes additional information beyond coordinates, such as the porous medium of each point in the Darcy dataset.

If the compression is based solely on spatial coordinates, \( \bm{H}_N \) only contains coordinate information; otherwise, when full point-wise information is used, \( \bm{H}_N = \bm{\hat{H}}_N \).
Whether \( \bm{H}_N \) is equal to \( \bm{\hat{H}}_N \) is an implementation detail and does not materially affect our discussion. Therefore, we can construct equivalent feature mappings $\varphi(\bm{Q})$, $\psi(\bm{K})$, $\bm{V}$ and $\mathcal{G}$ as in Eq.\eqref{eq:cross_vs_Linear},
\begin{equation}
    \begin{aligned}
        &\varphi(\bm{Q}) = \text{Softmax}(\text{Linear}^{(Q)}(\bm{H}_N)) \\
        &\psi^\top(\bm{K}) = \text{Softmax}(\text{Linear}^{(K)}(\bm{H}_N)^\top) \\
        &\bm{V}=\text{Linear}^{(V)}(\bm{\hat{H}}_N)\\
        &\mathcal{G}=W'_v\circ\text{Self-Attention}\circ \dots \circ \text{Self-Attention}
        \label{eq:cross_vs_Linear}
    \end{aligned}
\end{equation}

Overall, cross-attention with learnable tokens can also be regarded as a special case of linear attention.

\section{G. Visualizations}

In this section, we provide supplementary visualizations on weight maps and error maps. 

\begin{figure*}
    \centering
    \includegraphics[width=\linewidth]{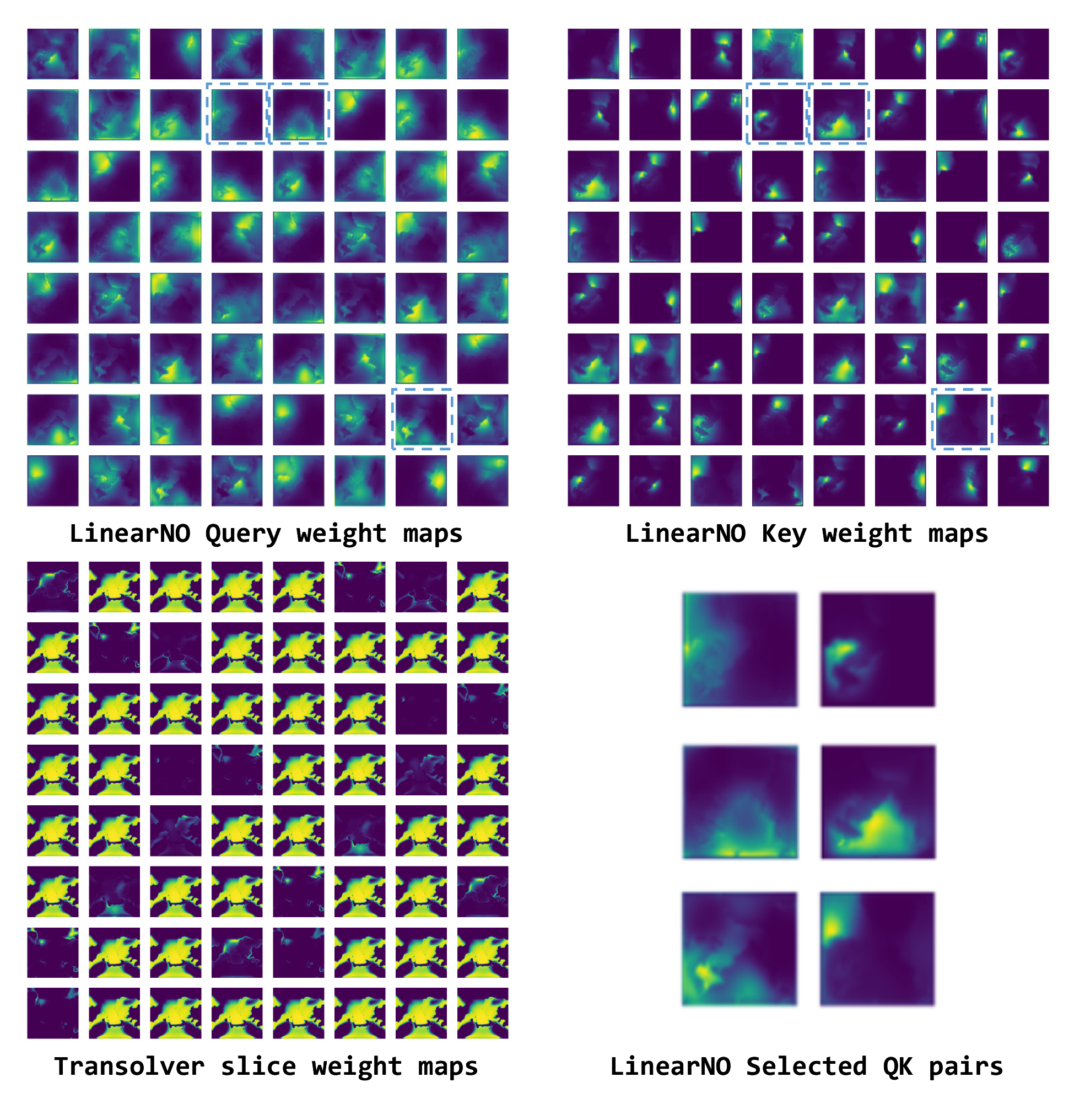}
    \caption{Visualization of weight maps on Darcy dataset.}
    \label{fig:darcy-qk}
\end{figure*}

\begin{figure*}
    \centering
    \includegraphics[width=\linewidth]{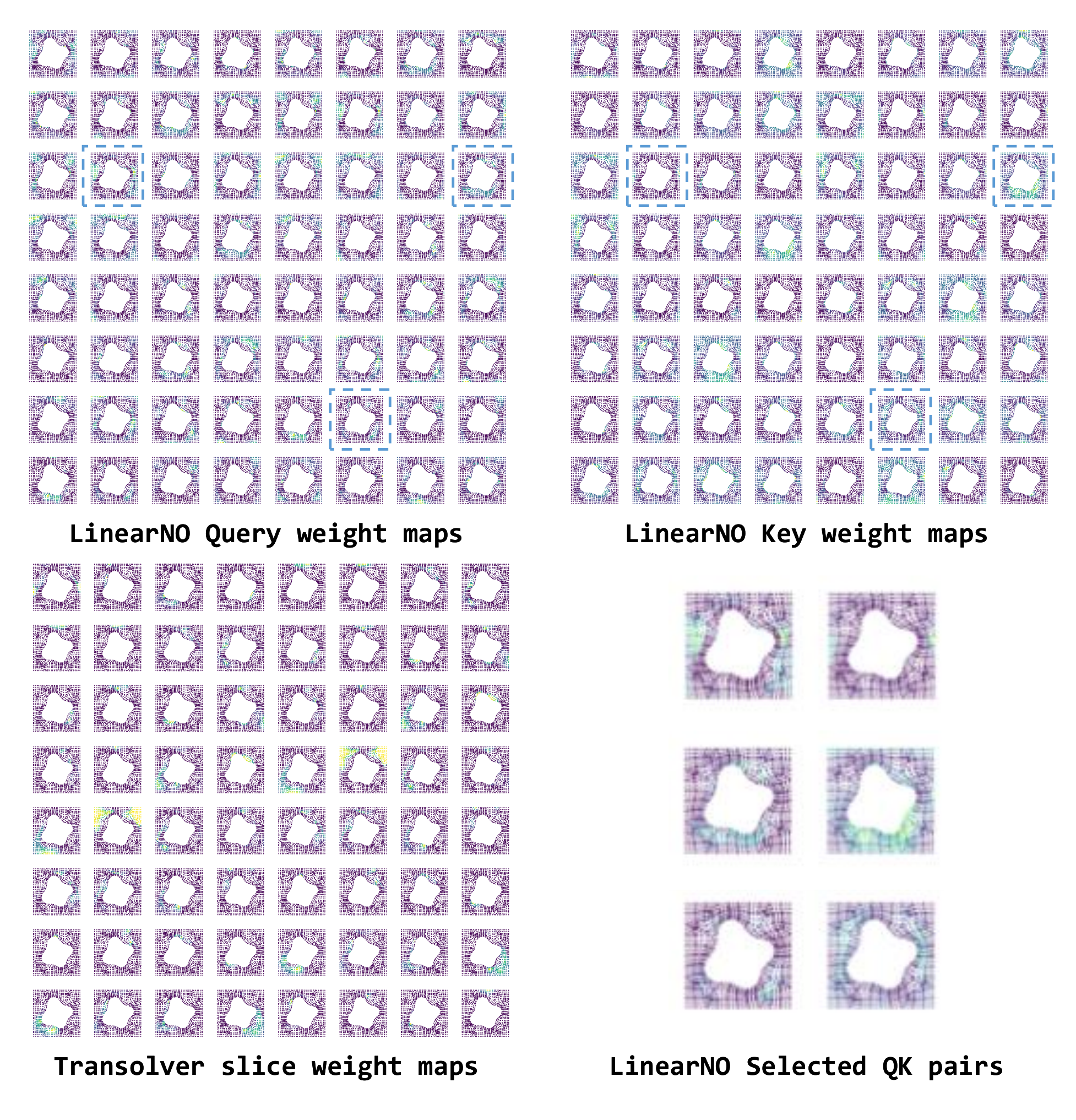}
    \caption{Visualization of weight maps on Elesticity dataset.}
    \label{fig:elas-qk}
\end{figure*}

\begin{figure*}
    \centering
    \includegraphics[width=\linewidth]{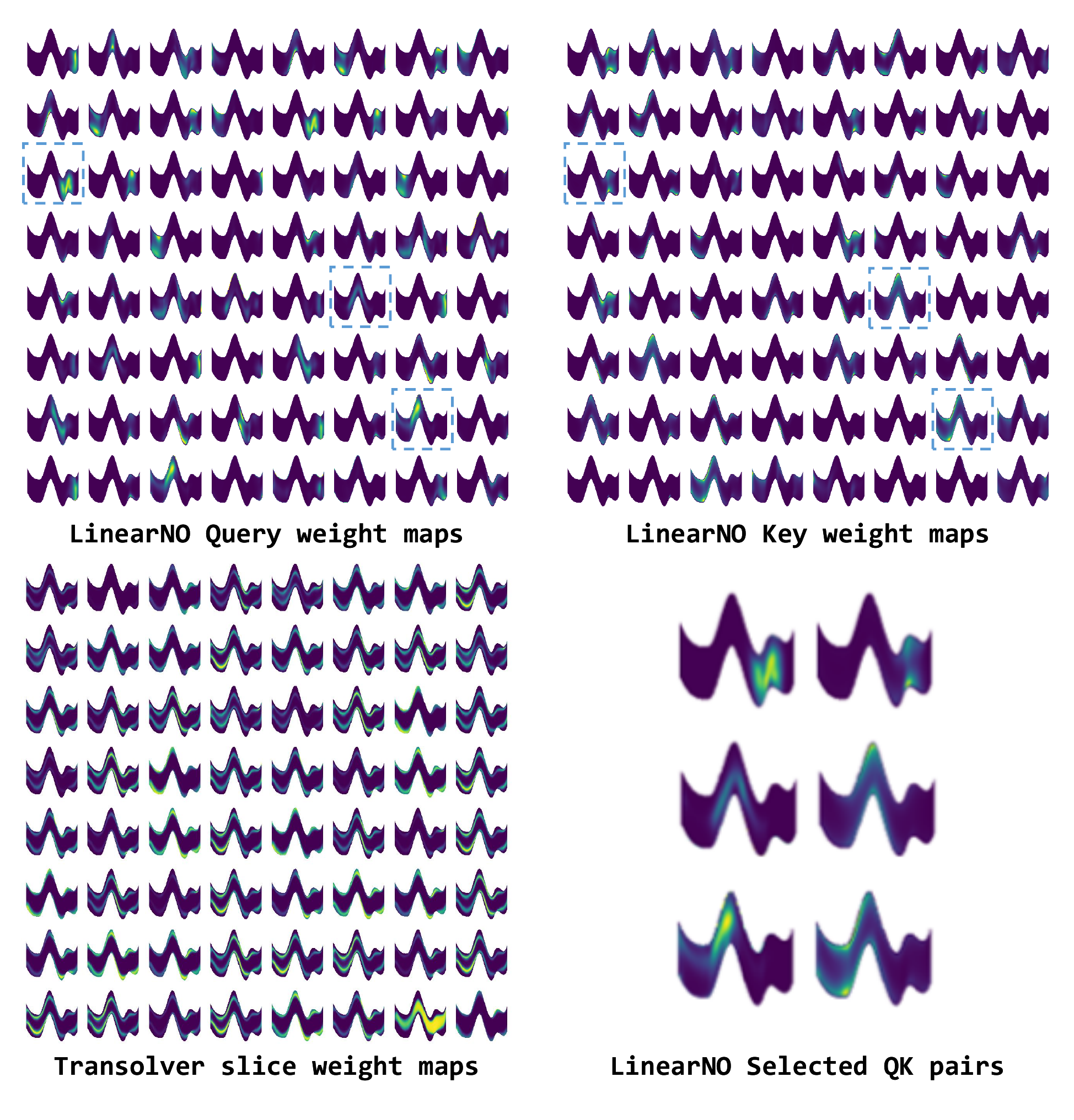}
    \caption{Visualization of weight maps on Pipe dataset.}
    \label{fig:pipe-qk}
\end{figure*}

\noindent\textbf{Weight maps.} The visualization of weight maps on dataset Darcy, Elasticity and Pipe are displayed in Fig.~\ref{fig:darcy-qk}, Fig.~\ref{fig:elas-qk} and Fig.~\ref{fig:pipe-qk}, respectively. Compared to Transolver, our LinearNO demonstrate more diverse patterns on attention maps. In addition, Query and Key in LinearNO show different attention patterns, which shows the necessity of asymmetric design.

\begin{figure*}
    \centering
    \includegraphics[width=0.9\linewidth]{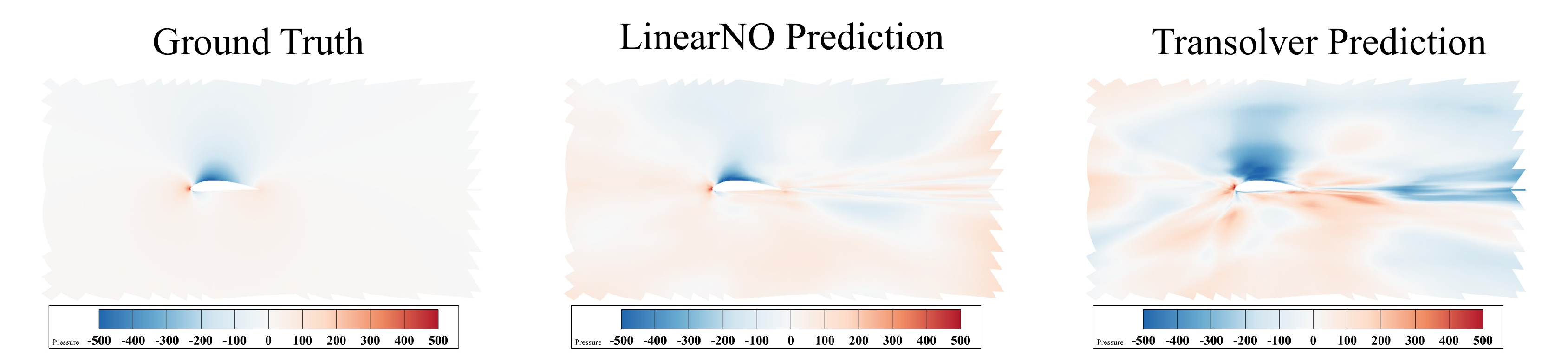}
    \caption{Prediction error on AirfRANS dataset.}
    \label{fig:airframs_error}
\end{figure*}

\begin{figure*}
    \centering
    \includegraphics[clip,trim=1mm 1mm 1mm 1mm, width=.9\linewidth]{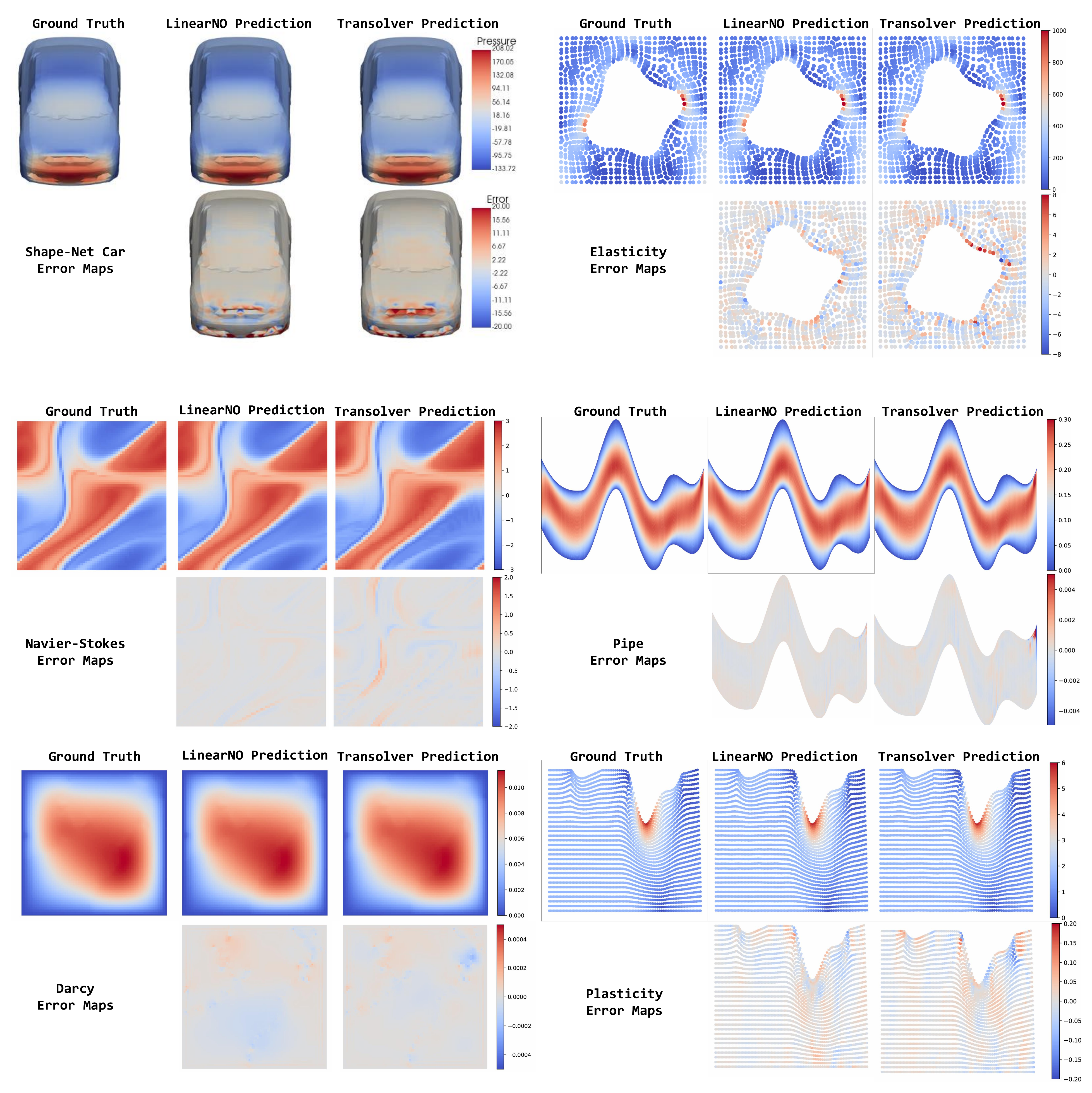}
    \caption{Example error maps on various datasets.}
    \label{fig:error-maps}
\end{figure*}

\noindent\textbf{Error maps.} Some example error maps for multiple datasets are illustrated in Fig.~\ref{fig:airframs_error} and Fig.~\ref{fig:error-maps}. The error maps show that LinearNO demonstrates higher prediction accuracy than Transolver in key areas where the flow field changes rapidly.


\



\ifreproStandalone
\end{document} 
\fi
\end{document}